\documentclass[letterpaper]{article} 
\usepackage{aaai2026}  
\usepackage{times}  
\usepackage{helvet}  
\usepackage{courier}  
\usepackage[hyphens]{url}  
\usepackage{graphicx} 
\usepackage{natbib}  
\usepackage{caption} 
\usepackage{algorithm}
\usepackage{algorithmic}

\usepackage{multirow}
\usepackage{multicol}
\usepackage[table]{xcolor} 
\usepackage{pifont}
\usepackage{booktabs}
\usepackage{amsmath}
\usepackage{makecell}
\usepackage{newfloat}
\usepackage{listings}
\DeclareCaptionStyle{ruled}{labelfont=normalfont,labelsep=colon,strut=off} 
\floatstyle{ruled}
\newfloat{listing}{tb}{lst}{}
\floatname{listing}{Listing}
\title{Zooming In on Fakes: A Novel Dataset for Localized AI-Generated Image Detection with Forgery Amplification Approach}
\author{
    Lvpan Cai\textsuperscript{\rm 1}\equalcontrib,
    Haowei Wang\textsuperscript{\rm 2}\equalcontrib,
    Jiayi Ji\textsuperscript{\rm 1}\equalcorr,
    Yanshu Zhoumen\textsuperscript{\rm 1}, \\
    Shen Chen\textsuperscript{\rm 2},
    Taiping Yao\textsuperscript{\rm 2},
    Xiaoshuai Sun\textsuperscript{\rm 1}\equalcorr
}
\affiliations{
    \textsuperscript{\rm 1}Key Laboratory of Multimedia Trusted Perception and Efficient Computing, Ministry of Education of China, School of Informatics, Xiamen University, Xiamen 361005, China\\  
    \textsuperscript{\rm 2}Youtu Lab, Tencent, China  
}

\usepackage{bibentry}

\begin{document}

\maketitle

\begin{abstract}
The rise of AI-generated image tools has made localized forgeries increasingly realistic, posing challenges for visual content integrity.
Although recent efforts have explored localized AIGC detection, existing datasets predominantly focus on object-level forgeries while overlooking broader scene edits in regions such as sky or ground.
To address these limitations, we introduce \textbf{BR-Gen}, a large-scale dataset of 150,000 locally forged images with diverse scene-aware annotations, which are based on semantic calibration to ensure high-quality samples. 
BR-Gen is constructed through a fully automated ``Perception-Creation-Evaluation'' pipeline to ensure semantic coherence and visual realism.
In addition, we further propose \textbf{NFA-ViT}, a Noise-guided Forgery Amplification Vision Transformer that enhances the detection of localized forgeries by amplifying subtle forgery-related features across the entire image. 
NFA-ViT mines heterogeneous regions in images, \emph{i.e.}, potential edited areas, by noise fingerprints. Subsequently, attention mechanism is introduced to compel the interaction between normal and abnormal features, thereby propagating the traces throughout the entire image, allowing subtle forgeries to influence a broader context and improving overall detection robustness.
Extensive experiments demonstrate that BR-Gen constructs entirely new scenarios that are not covered by existing methods.
Take a step further, NFA-ViT outperforms existing methods on BR-Gen and generalizes well across current benchmarks.
\end{abstract}

\begin{links}
    \link{Code}{https://github.com/clpbc/BR-Gen}.
\end{links}

\section{Introduction}
\label{sec:intro}

The rapid advancement of generative models, such as Generative Adversarial Networks (GANs)~\cite{DBLP:conf/iclr/KarrasALL18} and Diffusion Models (DMs)~\cite{dangdiff}, enable fine-grained image modifications through simple user interactions like masks, and prompts~\cite{ju2024brushnet}.
These techniques raise serious concerns about image authenticity~\cite{ferreira2020review}, especially on social media platforms, while democratizing the creation of creative content.
Consequently, the ability to detect whether visual content has been altered is becoming increasingly critical.

Around the AI-generated content (AIGC), a few early researches construct various datasets~\cite{wang2023dire,zhu2023genimage} and benchmarks~\cite{wang2020cnn,ojha2023towards} to improve the detection performance on fully synthesized images. 
However, these efforts often overlook localized generation scenarios, where only specific regions are modified. Although some recent works~\cite{guillaro2023trufor,he2025survey} have attempted to localize manipulations, their progress is constrained by limitations in existing datasets and detection paradigm.

\begin{figure}[]
    \centering
    \includegraphics[width=0.48 \textwidth]{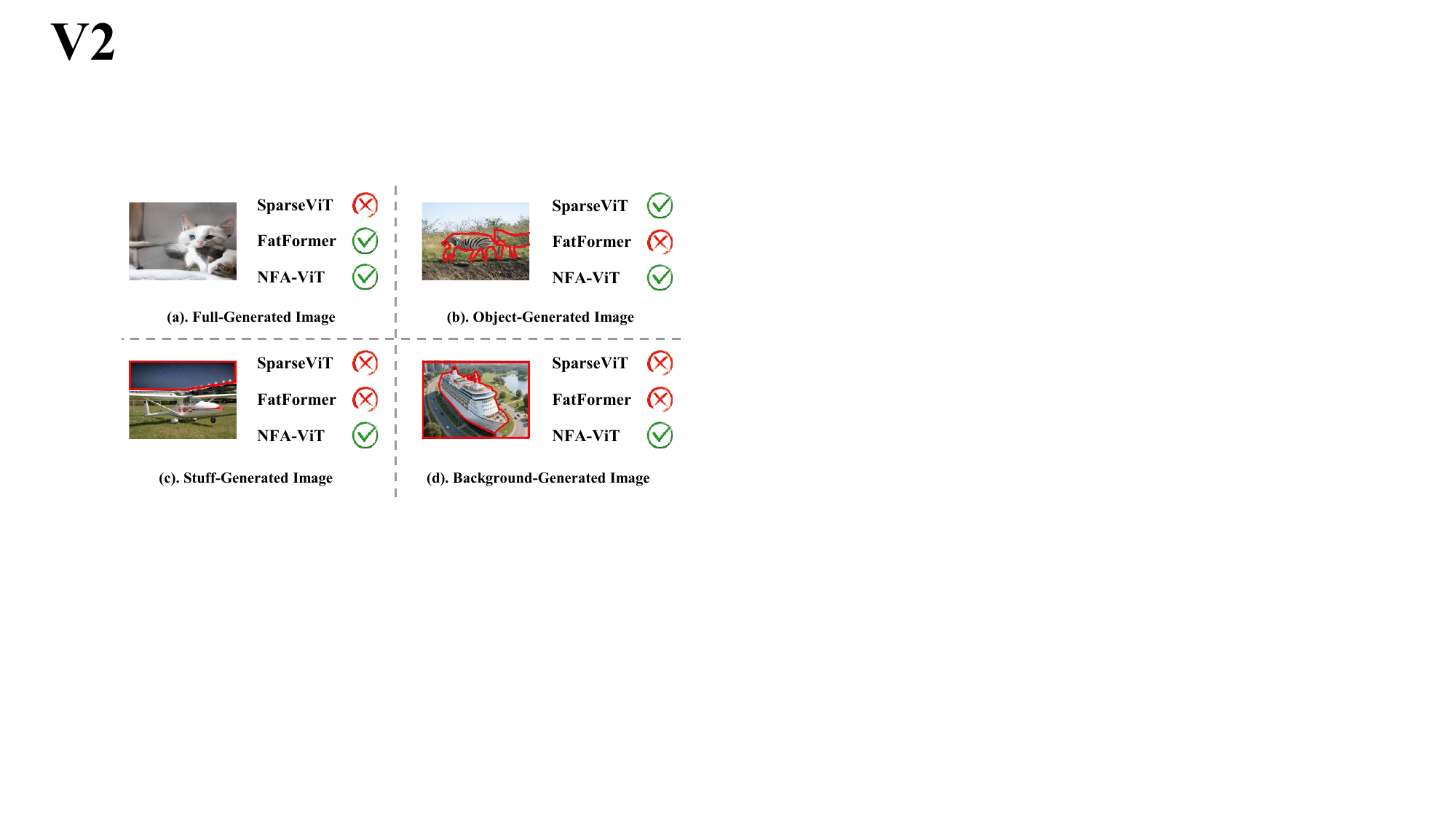}
    \caption{Comparison of four forgery scenarios in existing datasets. They mainly cover full-generated images and object-level forgeries, while forgeries in stuff and background regions remain largely unaddressed. Red regions show ground-truth forgeries. State-of-the-art models (FatFormer~\cite{liu2024forgery} and SparseViT~\cite{su2024can}) struggle with these new cases. Our proposed NFA-ViT achieves robust detection across all four scenarios. 
    The source data comes from open-source datasets GRE~\cite{sun2024rethinking}, COCO~\cite{lin2014microsoft}, and ImageNet~\cite{deng2009imagenet}.
    }
    
    \label{fig:motivation}
\end{figure}

\begin{table*}[t] 
  \centering

  \resizebox{1\textwidth}{!}{
  \begin{tabular}{lccccccc}
    \toprule
    \multirow{2}{*}{\textbf{Dataset}} & \multicolumn{2}{c}{\textbf{Dataset Scale}} & \multicolumn{2}{c}{\textbf{Gen. Category}} & \multicolumn{2}{c}{\textbf{Mask Type}} & \multirow{2}{*}{\textbf{Gen. Area}} \\
    \cmidrule(lr){2-3} 
    \cmidrule(lr){4-5}
    \cmidrule(lr){6-7}
     & {Real Images} & {Gen. Images} & {GAN-based} & {DM-based} & {Stuff} & {Background}  & \\ 
    \midrule 
    NIST16 ~\cite{guan2019mfc} & 0 & 564 & 1 & -  & \ding{55} & \ding{55} & Small \\ 
    DEFACTO ~\cite{mahfoudi2019defacto} & - & 149,000 & 1 & -  & \ding{55} & \ding{55} & Small \\ 
    IMD2020 ~\cite{novozamsky2020imd2020} & 35,000 & 35,000 & 1 & -  & \ding{55} & \ding{55} & Small \\ 
    CocoGLIDE ~\cite{guillaro2023trufor} & 512 & 512 & - & 1  & \ding{55} & \ding{55}& Small \\ 
    AutoSplice ~\cite{jia2023autosplice} & 2,273 & 3,621 & - & 1 & \ding{55} & \ding{55} & Small/Medium \\ 
    TGIF ~\cite{mareen2024tgif} & 3,124 & 74,976 & - & 2  & \ding{55} & \ding{55}& Small \\ 
    GRE ~\cite{sun2024rethinking} & - & 228,650 & 2 & 3  & \ding{55} & \ding{55} & Small \\ 
    SID-Set ~\cite{huang2025sidasocialmediaimage}& 70,000 & 140,000& - & 2 & \ding{55} & \ding{55} & Small/Full\\   
    \midrule
    \rowcolor{green!12} 
    \textbf{BR-Gen} (Ours) & 15,000 & 150,000 & 2 & 3& \ding{51} & \ding{51}  & Small/Medium/Large \\ 
    
    \bottomrule
  \end{tabular}
  }

  \caption{Summary of the attributes for localized AIGC detection datasets. Numbers indicate category quantities, and ``Area'' denotes the distribution range of forged regions.}
    \label{tab:dataset_attribute}
\end{table*}

Specifically, current localized AIGC datasets~\cite{guillaro2023trufor} suffer from two major limitations. (1) \textbf{Pervasive forgery region bias}. Previous datasets focus on salient objects or synthetic rectangular patches while defining forged areas, neglecting complex scene-level elements like sky, ground, vegetation, or structural background. 
Detectors trained on such data tend to overfit to object-centric artifacts and fail to generalize to more subtle or spatially distributed forgeries.
(2) \textbf{Uncontrollable editing quality}, which further limits their effectiveness. Many generated samples exhibit unrealistic textures, compression artifacts, or visible boundary seams due to low-quality generation pipelines and a lack of quality control. These flaws not only reduce visual plausibility but also make detection easier, masking the true difficulty of localized forgery detection in real-world scenarios. Fig.~\ref{fig:motivation} illustrates how data limitations affect model detection. The model fails to detect when faced with out-of-distribution data or complex scene-level elements.

To address these challenges, we present the Broader Region Generation (\textbf{BR-Gen}) dataset, a large-scale and high-quality benchmark containing 150,000 locally forged images with diverse region coverage. BR-Gen targets underrepresented ``stuff'' and ``background'' categories—including sky, ground, wall, grass, and vegetation—substantially broadening the scope of localized forgeries beyond objects. The dataset is constructed through a fully automated ``Perception-Creation-Evaluation'' pipeline that ensures semantic integrity and visual realism. Specifically, we use grounding and segmentation models to guide localized editing~\cite{ravi2024sam2,liu2023grounding}, diffusion-based generative models for content synthesis~\cite{podell2023sdxl, zhuang2024task,wang2025training}, and multi-stage perceptual evaluation metrics~\cite{fu2023dreamsim,radford2021learning} to validate image quality. Compared to prior datasets, BR-Gen offers broader region diversity, more realistic forgeries, and stronger alignment with real-world editing patterns.

Despite recent advances~\cite{su2024can,guillaro2023trufor,liu2022pscc} in localized forgery detection, existing methods exhibit two major limitations when evaluated on BR-Gen. First, they tend to detect generic discrepancies between regions without the capability to identify which parts are truly forged. This often leads to overfitting to dataset-specific biases, such as the typical size, shape, or location of forged areas, and results in significant errors when these patterns vary. See Case 3 in Fig.\ref{fig:seg_result} for an example. Second, when forgeries are small or embedded in visually inconspicuous regions (e.g., within complex backgrounds or ``stuff'' areas), the manipulation signals become extremely weak and are easily overshadowed by surrounding authentic content, rendering them difficult to detect. This issue is illustrated in Case 4 of Fig.\ref{fig:seg_result}.

To address these challenges, we introduce \textbf{NFA-ViT}, a novel architecture designed for robust localized forgery detection. NFA-ViT introduces a \textit{forgery amplification} mechanism that enhances weak manipulation cues through a dual-branch framework. Specifically, a dedicated noise fingerprint branch extracts subtle discrepancies between authentic and forged regions, while a visual transformer backbone incorporates these signals via noise-guided attention queries. This allows authentic regions to absorb and propagate discriminative forgery features across the image, thereby amplifying weak signals and improving detection sensitivity, particularly in sparse or spatially diffuse cases. Moreover, when forgery cues are diffused across the entire image, the classifier is no longer constrained to detecting discrepancies between isolated regions. Instead, it learns to distinguish globally between authentic and manipulated content based on the presence or absence of generative attributes. This adaptive learning paradigm significantly improves the model's ability to localize forgeries, regardless of their size, shape, or location. Extensive experiments on BR-Gen and other benchmarks demonstrate that NFA-ViT not only achieves state-of-the-art performance but also exhibits strong cross-dataset generalization, establishing a new foundation for scene-aware, fine-grained forgery detection.

In summary, our contributions are three-fold:
\begin{itemize} 
\item We identify key limitations in existing localized AIGC datasets, including region bias and low visual quality, and introduce BR-Gen, a large-scale benchmark with diverse and realistic scene-level forgeries. 
\item We propose NFA-ViT, a noise-guided forgery amplification transformer that leverages a dual-branch architecture to diffuse forgery cues into real regions through modulated self-attention, significantly improving the detectability of small or spatially subtle forgeries.
\item We conduct extensive experiments showing that BR-Gen is a more challenging datasets, and that NFA-ViT achieves strong detection and localization performance.
\end{itemize}

\begin{figure*}[]
    \centering
    \includegraphics[width=1 \textwidth]{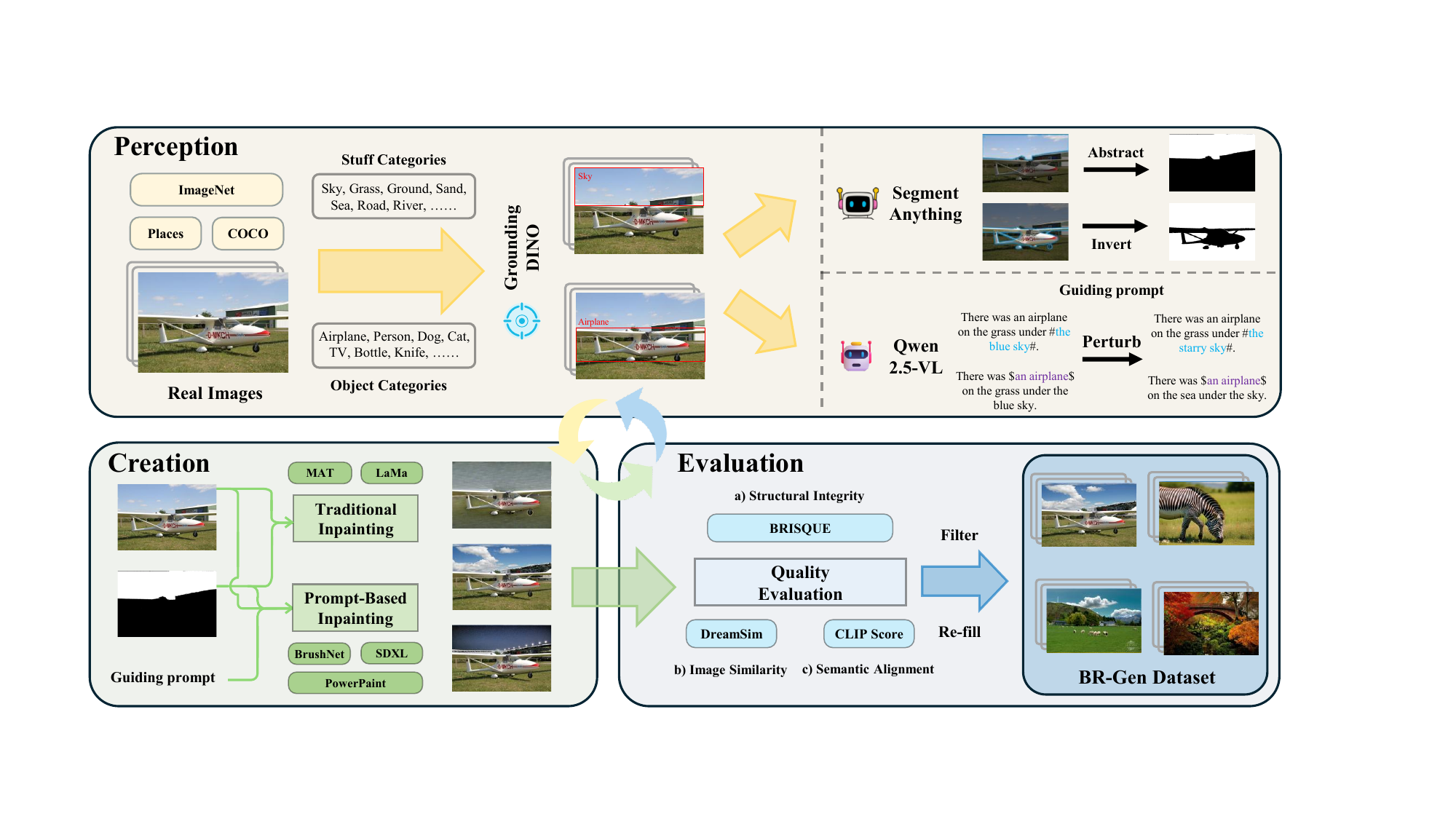}
    \caption{The automated pipeline for the BR-Gen dataset consists of three iterative stages: \textbf{Perception}, \textbf{Creation}, and \textbf{Evaluation}. These stages are applied to produce high-quality localized generation datasets through progressive refinement. All samples are sourced from publicly available datasets~\cite{zhou2017places, lin2014microsoft, deng2009imagenet}.}
    \label{fig:br-gen}
\end{figure*}

\section{Related Work}  
\label{sec:relatedwork}

\subsection{Generation Datasets}  
In recent years, Artificial Intelligence Generated Content (AIGC) has become widely used. Datasets~\cite{wang2020cnn,ojha2023towards} containing both real and generated images have been organized for training and evaluating detection systems. CNNSpot~\cite{wang2020cnn} uses GAN-generated images~\cite{DBLP:conf/iclr/KarrasALL18,chen2024decoupled}, and Chameleon~\cite{yan2024sanity} provides highly realistic test cases. However, these support only image-level tasks. For localized detection, existing datasets (Tab.~\ref{tab:dataset_attribute}) often use object masks from COCO~\cite{lin2014microsoft} or SAM~\cite{kirillov2023segment}, focusing on countable objects. This ignores large regions like sky or background, leading to bias and poor model generalization on such areas.  

\subsection{Generation Detection}  
The need for detecting AI-generated content has existed since the rise of deep learning. Early detection used spatial cues like color and reflection~\cite{mccloskey2018detecting,o2012exposing}. As GANs improved, methods like CNNSpot~\cite{wang2020cnn} used data augmentation for better generalization~\cite{karras2017progressive,chen2025dual,yang2025all,lin2025seeing}. Recent works exploit frequency or reconstruction patterns~\cite{wang2023dire,qian2020thinking}, but still focus on whole images. For localization, ManTra-Net~\cite{wu2019mantra} uses LSTM analysis, Trufor~\cite{guillaro2023trufor} uses noise patterns, and SparseViT~\cite{su2024can} uses sparse attention for top performance. Yet, they fail on complex or non-object regions, highlighting the need for our approach.


\section{BR-Gen Dataset}
Recent datasets for detecting localized forgery based on generative models~\cite{sun2024rethinking,guillaro2023trufor} have emerged. 
To address the gaps in existing datasets, we have taken into account the the neglected local edits in Stuff and Background, proposing a high-quality, scene-based local generation dataset named the Broader Region Generation (\textbf{BR-Gen}). 
We propose an automated pipeline with open-source models~\cite{ravi2024sam2,liu2023grounding,DBLP:journals/corr/abs-2502-13923}, generating local edited images from unannotated ones. 
As shown in Tab.~\ref{tab:dataset_attribute}, BR-Gen takes into full account diverse generation methods and tampering areas, addressing the shortcomings in the types of previous datasets.

\subsection{Real Image Collection}
We sampled images from three large-scale datasets like previous works~\cite{jia2023autosplice}: ImageNet~\cite{deng2009imagenet}, COCO~\cite{lin2014microsoft}, and Places~\cite{zhou2017places}. These datasets provide diverse scenes with rich semantic content, enhancing the diversity of dimensions.

\subsection{Localized Generation Pipeline}
To simulate real-world image editing processes while maintaining content and semantic consistency, we designed an automated pipeline that integrates multiple open-source models~\cite{ravi2024sam2,liu2023grounding,DBLP:journals/corr/abs-2502-13923}, as illustrated in Fig.~\ref{fig:br-gen}. The pipeline comprises three stages: ``\textbf{Perception-Creation-Evaluation}''. Below, we briefly describe each stage, with detailed process descriptions and dataset statistics provided in the appendix.

\subsubsection{Preception.}
This stage identifies regions of interest in real images and extracts semantic information to guide subsequent creation stages.  We select the candidate categories by ``thing category'' and ``stuff category'' from COCO. GroundingDINO~\cite{liu2023grounding} detects object bounding boxes for these categories, which are then converted into masks by SAM2~\cite{ravi2024sam2}. ``Stuff'' masks are obtained directly, while ``background'' masks are derived by inverting ``thing'' masks. To balance category distribution, we control the number of instances per category.

For prompt-based inpainting, we use  Qwen2.5-VL~\cite{DBLP:journals/corr/abs-2502-13923} to generate image descriptions. Then, we apply Semantic Perturbation to increase semantic diversity: for ``stuff'', we replace content within ``\#'' symbols (e.g., ``the blue sky'' $\to$ ``the starry sky''); for ``background'', we replace text outside ``\$'' symbols. Extra semantically similar object lists ensure plausible edits.

\begin{figure}[]
    \centering
    \includegraphics[width=0.48 \textwidth]{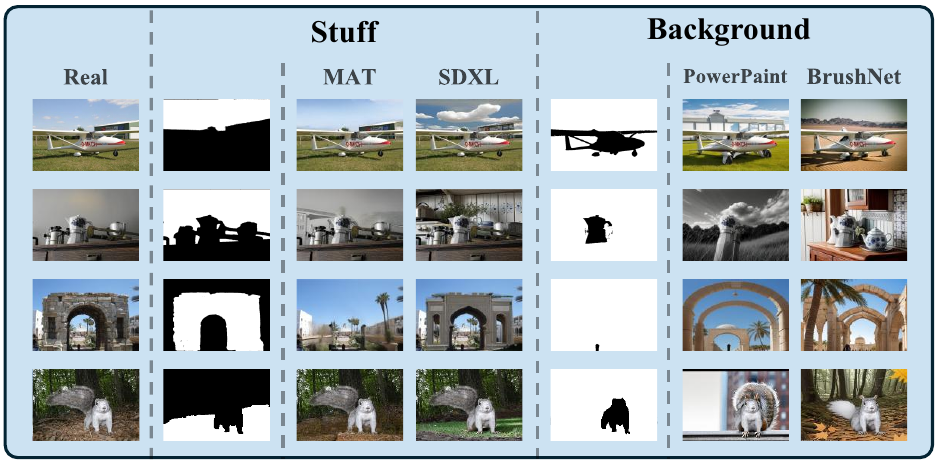}
    \caption{Partial examples from BR-Gen dataset. The real data comes from the open datasets Places~\cite{zhou2017places}, COCO~\cite{lin2014microsoft}, and ImageNet~\cite{deng2009imagenet}.}
    
    \label{fig:cases}
\end{figure}

\subsubsection{Creation.} 
This stage uses region masks and guiding prompts to generate locally forged images. We employ five diversity state-of-the-art inpainting methods: two GAN-based (LaMa~\cite{suvorov2022resolution}, MAT~\cite{li2022mat}) and three diffusion-based (SDXL~\cite{podell2023sdxl}, BrushNet~\cite{ju2024brushnet}, PowerPaint~\cite{zhuang2024task}). GAN methods require only the image and mask, while diffusion methods also require the guiding prompt.

\subsubsection{Evaluation.} 
This stage filters high-quality generated images using image quality assessment methods. Image quality is assessed based on:
(a) BRISQUE~\cite{mittal2012no} for structural integrity;
(b) DreamSim~\cite{fu2023dreamsim} for image similarity; and
(c) CLIP scores for prompt alignment. 
Low-quality samples are removed. Missing data is replenished through iterative generation to maintain dataset size.

\subsection{Showcase}
To visually illustrate the effectiveness of our automated pipeline and the quality of the resulting dataset, we present examples in Fig.~\ref{fig:cases}, covering diverse data sources, mask types, and inpainting methods. Additional detailed visualization samples are provided in the appendix.

\subsection{Dataset Splits}
Following the standard dataset partitioning approach for localized detection, the dataset is randomly divided into training, validation, and test sets using an 8:1:1 ratio. This division is applied to the subset of real images within the dataset. Regardless of the data source (ImageNet, COCO, Places), the partitioning ratios remain consistent. As a result, the training set includes 12,000 real images, while both the validation and test sets each contain 1,500 real images.

To prevent data leakage that could compromise the evaluation of model performance on the dataset, the partitioning of region masks and generated image sets is synchronized with the pre-divided real fig/image set. Thus, each triplet (real image, mask, forged image) is assigned to a single dataset partition, ensuring data integrity.

\begin{figure*}[t]
    \centering
    \includegraphics[width=1 \textwidth]{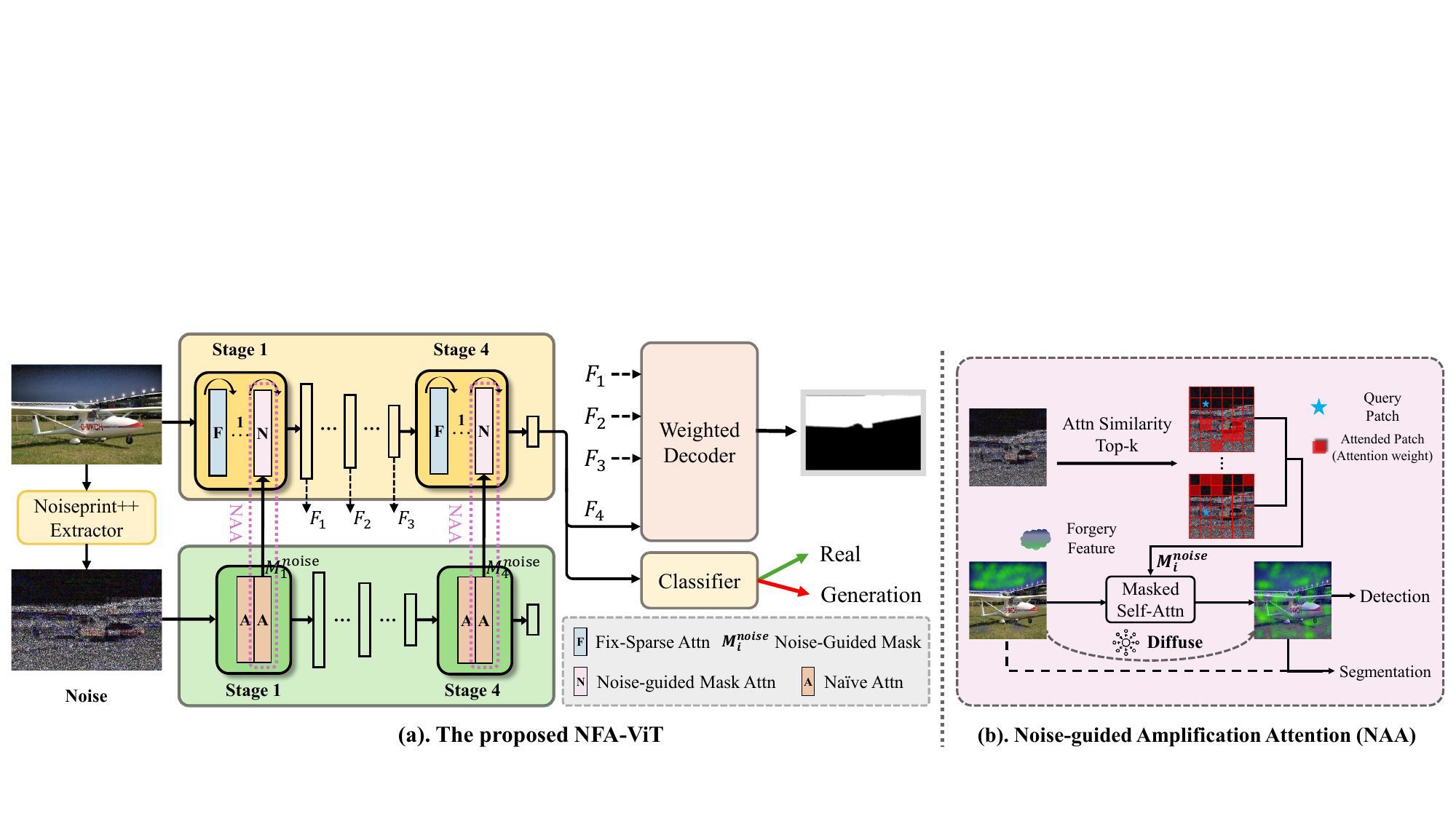}
    \caption{The proposed NFA-ViT framework, which contains dual branches of noise and image, uses noise information to guide the focus area of the image. For the image encoder, a sparse attention mechanism is introduced. }
    \label{fig:nas-vit}
\end{figure*}

\section{The proposed NFA-ViT}
\subsection{Overview}
To further enhance the performance of local AIGC detection, we propose the Noise-guided Forgery Amplification Vision Transformer (NFA-ViT) to take advantages of the non-homologous between the generated regions and real regions.
Noise fingerprints and related features exhibit discrepancies across different regions~\cite{guillaro2023trufor, wang2023dire}. However, related studies suffer from significant limitations, such as cannot determine which region is forged and being insensitive to subtle and tiny-area forgeries. We propose a novel detection mechanism called \textit{forgery amplification}. NFA-ViT leverages noise information as guidance to amplify and diffuse localized forgery features across the entire image, making forgery features more distinguishable while ensuring that the judgment of real images remains unaffected.

Fig.~\ref{fig:nas-vit}(a) illustrates the overall framework. For an input RGB image $x$, we first use the noise extractor Noiseprint++~\cite{guillaro2023trufor} to extract the noise trace $n$ of the image. We verify the effectiveness of the features and provide detailed information in the appendix.
Subsequently, $x$ and $n$ are jointly fed into a dual-branch network. For each stage, the Noise-guided Mask $M_i^n$ from the noise branch is used to guide the learning of the proposed \textbf{Noise-guided Amplification Attention} (NAA) in the image branch. With the NAA, real regions directly focus on the differential forgery regions, gradually diffusing forgery features into real regions:
\begin{equation}
    \begin{aligned}
P_{l+1}(i,j) = \alpha \cdot P_l(i,j) + \beta \cdot \frac{\sum_{(m,n)\in\mathcal{N}(i,j)} P_l(m,n)}{|\mathcal{N}(i,j)|} ,
    \end{aligned}
\end{equation}
where $P_{l}(i,j)$ represents the vector of real features at position $(i,j)$ in layer $l$, and $\mathcal{N}(i,j)$ represents the set of neighboring forgery features. Through layer-by-layer diffusion operation, forgery features expand from local areas to global areas.

Meanwhile, residual connections in the Noise-guided Mask Attnention maintain the original image features.
Finally, outputs from all stages are fed into a light-weight weighted decoder to generate the final results.

\subsection{Noise-guided Amplification Attention}
For simplicity, we describe the attention workflow for a single attention head.
We first introduce Fix-Sparse Attention~\cite{su2024can}, which helps refine localization by eliminating irrelevant semantic information. However, Fix-Sparse Attention disrupts semantics from a global perspective to learn non-semantic features, lacking the ability to aggregate and recognize local information.

Based on it, the proposed Noise-guided Amplification Attention (NAA) using noise signals to guide the amplification of forged features in images. Specifically, each stage of the noise branch is composed with vanilla self-attention. In the last layer of the vanilla attention, we take the noise as query matrix $Q^{noise}$ and key matrix $K^{noise}$ to compute the attention matrix $A^{noise}$ as follows:  
\begin{equation}
    \begin{aligned}
A^{noise}=\text{Softmax}(\frac{{Q^{noise}K^{noise}}^T}{\sqrt{d}}).
    \end{aligned}
\end{equation}

To amplify and diffuse features of forged regions toward real regions, we identify the $k$ most dissimilar $K$ values corresponding to each $Q$ in $A^{noise}$, forming a Noise-guided Mask $M^{noise}$, which represents the forged regions corresponding to real regions:  
\begin{equation}
    \begin{aligned}
M^{noise}=\mathbf{1}\left[ \text{Top-}k(-A^{noise}) \right].
    \end{aligned}
\end{equation}

Since the number of heads in corresponding layers of the two branches is identical, it is feasible to use noise information to guide image processing. The mask $M^{noise}$ is then inserted into the last layer in the each stage of image branch. Taking image feature as $Q^{image}$ and $K^{image}$, the output features $F$ is:
\begin{equation}
    \begin{aligned}
F_{ij} = \text{Softmax}\left( \frac{{Q^{image}K^{image}}^T}{\sqrt{d}} \right)_{ij} \text{iff} \quad M^{noise}_{ij} = 1,
    \end{aligned}
\end{equation} 
where $i$ and $j$ are the pixel location.
In this way, real-region features learn from generated regions, integrating traces of forgery.

\begin{table*}[]

  \centering
  \resizebox{\textwidth}{!}{
    \begin{tabular}{@{}clccccccccc@{}}
    \toprule
     \multirow{2}{*}{\textbf{Task}} & \multirow{2}{*}{\textbf{Method}} &\multirow{2}{*}{\textbf{Real Recall@50}} &
     \multicolumn{4}{c}{\textbf{BR-Gen dataset}} & \multicolumn{2}{c}{\textbf{Split A}} & \multicolumn{2}{c}{\textbf{Split B}} \\
        \cmidrule(lr){4-7}
        \cmidrule(lr){8-9}
        \cmidrule(lr){10-11}
     
   ~ & ~&~ & \multicolumn{1}{c}{\textbf{F1}} & \multicolumn{1}{c}{\textbf{AUC}} & \multicolumn{1}{c}{\textbf{Recall@50}} & \multicolumn{1}{c}{\textbf{IoU}} & \multicolumn{1}{c}{\textbf{GAN R@50}} & \multicolumn{1}{c}{\textbf{Diffusion R@50}} & \multicolumn{1}{c}{\textbf{Background R@50}} & \multicolumn{1}{c}{\textbf{Stuff R@50}} \\ \midrule
    \multirow{5}{*}{\textbf{\makecell{Localized\\Detection}}} & ManTranet~\cite{wu2019mantra} & 0.822 & 0.123 & - & 0.069 & 0.008 \textcolor{red}{($\downarrow$ 0.133)} & 0.074 & 0.061 & 0.077 & 0.058 \\
    & MVSS-Net~\cite{dong2022mvss} & 0.862 & 0.183 & 0.344 & 0.122 & 0.029 \textcolor{red}{($\downarrow$ 0.424)}& 0.154 & 0.098 & 0.154 & 0.092 \\
    & PSCC-Net~\cite{liu2022pscc} & 0.806 & 0.253 & 0.284 & 0.164 & \textbf{0.052} \textcolor{red}{($\downarrow$ 0.426)}& 0.166 & 0.161 & 0.170 & 0.155 \\
    & Trufor~\cite{guillaro2023trufor} & 0.881 & 0.295 & 0.319 & 0.194 & 0.048 \textcolor{red}{($\downarrow$ 0.630)}& 0.195 & 0.194 & 0.199 & 0.187 \\
    & SparseViT~\cite{su2024can} & 0.735 & 0.277 & - & 0.203 & \underline{0.049} \textcolor{red}{($\downarrow$ 0.649)} & 0.214 & 0.186 & 0.205 & 0.202 \\ 
    \midrule
    
    \multirow{4}{*}{\textbf{\makecell{AIGC\\Detection}}} &  LGrad~\cite{tan2023learning} & \underline{0.974} & 0.165 & \textbf{0.635} & 0.088 & - & 0.101 \textcolor{red}{($\downarrow$ 0.762)}& 0.057 & 0.093 & 0.085 \\
    & FreqNet~\cite{DBLP:conf/aaai/Tan0WGLW24} & 0.767 & 0.360 & 0.472 & 0.231 & - & 0.244 \textcolor{red}{($\downarrow$ 0.671)}& 0.228 & 0.240 & 0.229 \\
    & NPR~\cite{tan2024rethinking} & 0.894 & \underline{0.443} & 0.501 & \underline{0.300} & - & \underline{0.323} \textcolor{red}{($\downarrow$ 0.602)}& \underline{0.290} \textcolor{red}{($\downarrow$ 0.662)}& \underline{0.318} & 0.289 \\
    & FatFormer~\cite{liu2024forgery} & \textbf{0.989} & \textbf{0.493} & \underline{0.606} & \textbf{0.331} & - & \textbf{0.358} \textcolor{red}{($\downarrow$ 0.626)}& \textbf{0.321} \textcolor{red}{($\downarrow$ 0.629)}& \textbf{0.349} & \underline{0.310} \\
    \bottomrule
    \end{tabular}
  }

    \caption{The cross-domain results on BR-Gen. The evaluation methods include AIGC detection and localization detection. We \textbf{bold} the best result and mark the second-best result with an \underline{underline}. The red decline values indicate the level of performance decrease compared to the original dataset~\cite{ma2025imdl,liu2024forgery,tan2024rethinking}. Since some methods don't provide corresponding indicators, some values are missing.}
      \label{tab:cross_domain}
\end{table*}

\begin{table*}[t]


  \centering
  \resizebox{\textwidth}{!}{
    \begin{tabular}{@{}clccccccccc@{}}
    \toprule
     \multirow{2}{*}{\textbf{Task}} & \multirow{2}{*}{\textbf{Method}} &\multirow{2}{*}{\textbf{Real Recall@50}} &
     \multicolumn{4}{c}{\textbf{BR-Gen dataset}} & \multicolumn{2}{c}{\textbf{Split A}} & \multicolumn{2}{c}{\textbf{Split B}} \\
        \cmidrule(lr){4-7}
        \cmidrule(lr){8-9}
        \cmidrule(lr){10-11}
     
   ~ & ~&~ & \multicolumn{1}{c}{\textbf{F1}} & \multicolumn{1}{c}{\textbf{AUC}} & \multicolumn{1}{c}{\textbf{Recall@50}} & \multicolumn{1}{c}{\textbf{IoU}} & \multicolumn{1}{c}{\textbf{GAN R@50}} & \multicolumn{1}{c}{\textbf{Diffusion R@50}} & \multicolumn{1}{c}{\textbf{Background R@50}} & \multicolumn{1}{c}{\textbf{Stuff R@50}} \\ \midrule
    \multirow{5}{*}{\textbf{\makecell{Localized\\Detection}}} & ManTranet~\cite{wu2019mantra} & 0.853 & 0.774 & - & 0.760 & 0.632 & 0.772 & 0.760 & 0.778 & 0.754  \\
    & MVSS-Net~\cite{dong2022mvss} & 0.903 & 0.892 & 0.924 & 0.883 & 0.671 & 0.913 & 0.846 & 0.889 & 0.856  \\
    & PSCC-Net~\cite{liu2022pscc} & 0.935 & 0.894 & 0.937 & 0.861 & 0.705 & 0.898 & 0.840 & 0.867 & 0.844  \\
    & Trufor~\cite{guillaro2023trufor} & 0.944 & 0.918 & 0.942 & 0.896 & 0.779 & 0.915 & 0.865 & 0.903 & 0.871  \\
    & SparseViT~\cite{su2024can} & 0.984 & 0.946 & - & 0.911 & \underline{0.824} & \underline{0.958} & 0.872 & 0.931 & 0.907  \\ 
    \midrule
    
    \multirow{6}{*}{\textbf{\makecell{AIGC\\Detection}}} &  LGrad~\cite{tan2023learning} & 0.937 & 0.831 & 0.872 & 0.755 & - & 0.801 & 0.732 & 0.775 & 0.738  \\
    & DIRE~\cite{wang2023dire} & 0.939 & 0.823 & 0.825 & 0.742 & - & 0.750 & 0.744 & 0.762 & 0.739  \\
    & FreqNet~\cite{DBLP:conf/aaai/Tan0WGLW24} & 0.825 & 0.699 & 0.702 & 0.631 & - & 0.659 & 0.614 & 0.648 & 0.622  \\
    & NPR~\cite{tan2024rethinking} & 0.946 & 0.922 & 0.933 & 0.902 & - & 0.938 & 0.884 & 0.921 & 0.893  \\
    & FatFormer~\cite{liu2024forgery} & \underline{0.990} & {0.961} & {0.971} & {0.935} & - & 0.955 & {0.913} & {0.949} & {0.915}  \\
    & AIDE~\cite{yan2024sanity} & 0.986 &\underline{0.964}& \underline{0.973} & \underline{0.941} & - & 0.950& \underline{0.932} & \textbf{0.965} & \underline{0.937}  \\
    \midrule
    \rowcolor{green!12}
    & \textbf{NFA-ViT} ~(ours) & \textbf{0.992} & \textbf{0.972} & \textbf{0.979} & \textbf{0.953} & \textbf{0.907} & \textbf{0.972} & \textbf{0.941} & \underline{0.961} & \textbf{0.948} \\
    
    \bottomrule
    \end{tabular}
  }

      \caption{The evaluation results of BR-Gen in-domain testing. After training the model on the BR-Gen training set, in-domain evaluation was conducted on the test set.}
        \label{tab:in_domain}
\end{table*}

\subsection{Weighted Decoder}
Current multi-level feature fusion methods often use addition or concatenation~\cite{lin2017feature}, producing feature maps through fixed linear aggregation without accounting for the varying contributions of hierarchical features to final maps. To improve region mask prediction, we propose a simple yet efficient decoder design. This approach introduces learnable scaling parameters $\gamma_i\ (1 \leq i \leq 4)$ for each hierarchical feature map, enabling adaptive weighted fusion by modulating layer-wise contributions.

The decoder processes four hierarchical features $F_i\ (1 \leq i \leq 4)$ extracted from the encoder. Each feature map $F_i$ is first projected uniformly to 512 channels using linear layers. Features $F_{2,3,4}$ are then upsampling to $\frac{1}{4}$ of the original resolution to align with the spatial dimensions of $F_1$. Each feature map $F_i$ is scaled by its corresponding parameter $\gamma_i$ before being summed. The aggregated features are compressed via linear projection to produce $\bar{M}$, which is subsequently upsampling to the original image resolution for mask prediction $\hat{M}$. The process is defined as follows:
\begin{equation}
    \begin{aligned}
\hat{F}_i=Upscale(MLP({F}_i)),\quad 1\leq i\leq4,
    \end{aligned}
\end{equation}
\begin{equation}
    \begin{aligned}
\hat{M}=Upscale(MLP(\sum_{i}^{4}({\hat{F}_i\times \gamma_i}))).
    \end{aligned}
\end{equation}

This design effectively balances multi-scale features by suppressing irrelevant information and emphasizing critical features via parameterized hierarchical integration.

\subsection{Loss Function}
For detection tasks, we use a lightweight backbone network~\cite{he2016deep} to extract features $F_4$, which generate predictions $\hat{y}$. For localization tasks, predicted masks $\hat{M}$ from the Weighted Decoder are utilized. Given ground truth labels $y$ and ground truth region masks $M$, the NFA-ViT model is trained using the following objective function:
\begin{equation}
    \begin{aligned}
        \mathcal{L} = \mathcal{L}_{\text{cls}}(y, \hat{y}) + \mathcal{L}_{\text{seg}}(M, \hat{M}),
    \end{aligned}
\end{equation}
where both $\mathcal{L}_{\text{cls}}$ and $\mathcal{L}_{\text{seg}}$ are binary cross-entropy loss.

\section{Experiment}
\subsection{Experimental Setup}
\noindent \textbf{Protocols and Evaluation Metrics.}
We conducted a comprehensive evaluation of the BR-Gen dataset using methods that include AIGC detection and local AIGC detection. First, we assessed the generalization ability of current models on BR-Gen. Next, we performed in-domain testing to evaluate model performance within the BR-Gen. 
Additionally, we tested the models on existing traditional benchmark. For fairness, the tampered images were downsampled to match the number of the authentic images when training. We used several metrics for a thorough evaluation from detection to localization: (1) Recall@50, classification metrics, which measures the model's ability to correctly identify categories; (2) F1 and AUC, classification metrics, are used to evaluate the overall performance and stability of the model; (3) IoU, localization metrics, which measures segmentation accuracy at the localization level.

\noindent \textbf{Implementation Details.}
For NFA-ViT, we use SegFormer~\cite{xie2021segformer} as the backbone, with the image and noise encoders being the b2 and b0 versions, respectively. During training, the model is optimized using the Adam optimizer~\cite{kinga2015method}, with an initial learning rate of $5\times 10^{-3}$ and a weight decay of $1\times 10^{-6}$. Using the Warmup and CosineAnnealing to help models achieve better convergence. In the Noise-Guided Amplification Attention mechanism, the $\text{Top-}k$ ratio is set to 25\%. Equal weights are given to all parts of the loss function. All experiments are run for 30 epochs with a batch size of 64.

\subsection{Experimental Results}

\noindent \textbf{Cross-domain on BR-Gen.}
To evaluate data bias in current generated image detection tasks, we conducted cross-domain testing on the BR-Gen dataset using detection models from two tasks: AIGC detection and local AIGC detection. We directly test the released trained models of those models, which are trained on data~\cite{novozamsky2020imd2020, karras2017progressive} that shares the same source as BR-Gen.
We divided the BR-Gen's into ``Split A'' and ``Split B'' according generation method and mask sources. Among them, ``Split A'' is categorized based on the generation method, specifically GAN and diffusion, while ``Split B'' is categorized based on the source of the masks, \emph{i.e.}, stuff and background.
For each model, we also compared its performance drop relative to its original report~\cite{ma2025imdl,liu2024forgery,tan2024rethinking},  under the corresponding data distribution. The experimental results are shown in Tab.~\ref{tab:cross_domain}.

The results show that local AIGC detection models perform worse overall compared to AIGC detection models. Meanwhile, all methods show a clear drop in generalization performance compared to their original reports. Although these models maintain high recall for real images, they show very low recall for partially generated content, indicating a consistent misclassification of tampered images. In terms of localization ability, the highest IoU value is only \textbf{0.052}, showing a broad failure to correctly identify tampered regions. These findings confirm the presence of data bias in current tasks and support the improved balance of our dataset, providing a strong base for improving detection performance in this field.

\begin{figure}[]
    \centering
    \includegraphics[width=0.47 \textwidth]{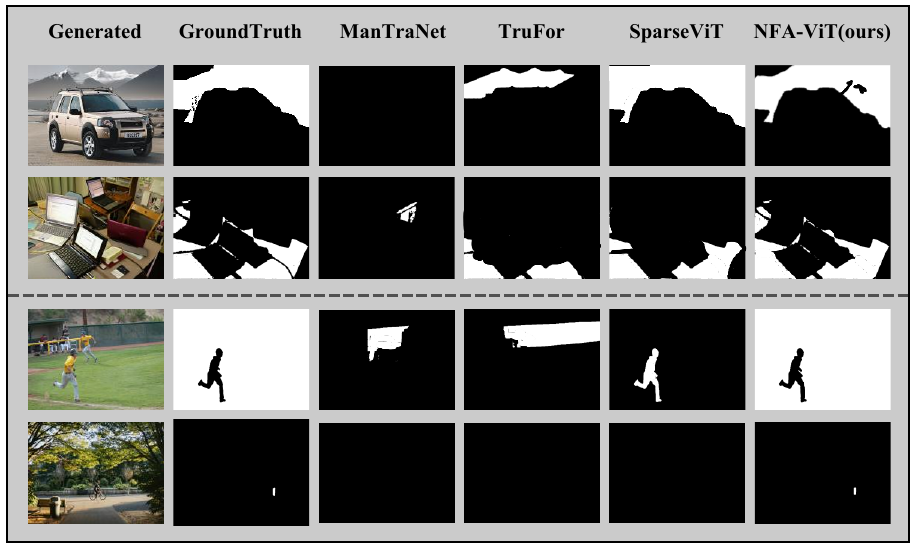}
    \caption{Localization results of different models. We compared images generated by two types of masks. All samples are sourced from publicly available datasets~\cite{zhou2017places, lin2014microsoft, deng2009imagenet}.}
    \label{fig:seg_result}
\end{figure}

\noindent \textbf{In-domain on BR-Gen.}
To evaluate performance on BR-Gen, we trained and tested detection models on the same dataset. Results, shown in Tab.~\ref{tab:in_domain}, indicate clear performance improvements post-training. For detection, AIDE achieved a recall rate of 94.1\% for generated images in the dataset, showing the complexity and difficulty of BR-Gen data during in-domain testing. For localization, SparseViT achieved an IoU of 82.4\%. A detailed analysis of dataset subtypes showed that in ``Split A'', GAN-based detection performed better than Diffusion-based methods. This is partly because GAN-generated images in the dataset were of lower quality and had more visible forgery features. In ``Split B'', background types were easier to detect than stuff types, and larger forged areas gave models more useful information.

 We carried out the same evaluation on NFA-ViT, and the results showed that NFA-ViT achieved better performance across all metrics. The F1 score reached \textbf{0.972}, surpassing AIDE by 0.8\%. For localization, the IoU reached \textbf{0.907}, outperforming SparseViT by 8.3\%, showing the value of amplifying forgery signals. Localization results from several models were visualized in Fig.~\ref{fig:seg_result}. By effectively addressing regional discrepancies and subtle forgeries, NFA-ViT achieves the best segmentation performance, while other models showed clear gaps.

\subsection{Ablation Studies}

\begin{table}[]
    \centering
        \resizebox{0.47\textwidth}{!}{
        \begin{tabular}{ccc|ccc}
            \toprule
            \textbf{Noise} & \textbf{NAA} & \textbf{\makecell{Weighted\\Decoder}} & \textbf{Mani R@50} & \textbf{Real R@50} & \textbf{IoU}\\
            \midrule
            \ding{55} & \ding{55} & \ding{55}& 0.880 & 0.925&0.759\\
            \ding{51} & \ding{55} & \ding{55}& 0.892 & 0.938&0.798\\

            \ding{51} & \ding{51} & \ding{55}& 0.937 & 0.972&0.833\\
            \ding{51} & \ding{55}  & \ding{51} & 0.913 & 0.959&0.867\\
            \rowcolor{green!12} 
            \ding{51} & \ding{51} & \ding{51}& \textbf{0.953} & \textbf{0.992}&\textbf{0.907} \\
            \bottomrule
        \end{tabular}}

        \caption{Components of the NFA-ViT.}
        
        \label{tab:ab_structure}

\end{table}

\noindent \textbf{Effectiveness of each component.}
We conducted structure ablations to systematically evaluate the contribution of each component within NFA-ViT and the results are presented in the Tab. ~\ref{tab:ab_structure}. We observed that each proposed component contributes positively to both detection and localization performance: The addition of the Noise branch improved localization by about 4\%, showing that noise differences between authentic and forged regions provide valuable cues. The inclusion of the NAA module further enhanced detection accuracy, demonstrating its efficacy in amplifying forgery traces across the entire image.And the weighted decoder significantly improved localization accuracy by dynamically adjusting the contribution of features from different layers, thereby refining the final localization output. Ultimately, combining all proposed components achieved the best overall performance.

\begin{table}[]
  
  \centering
  \resizebox{0.44\textwidth}{!}{
  
    \begin{tabular}{@{}cccc@{}}
    \toprule
      & \textbf{Gen. R@50} & \textbf{Real R@50} & \textbf{IoU} \\
     
    \midrule
    10\% & 0.945 & 0.989 & 0.887 \\
    \rowcolor{green!12}
    25\% & \textbf{0.953} & 0.992 & \textbf{0.907} \\ 
    50\% & 0.947 & \textbf{0.993} & 0.897 \\ 
    \bottomrule
    \end{tabular}}
  \caption{Different value of $\text{Top-}k$ in Noise-Guided Amplification Attention.}
  \label{tab:ab_top-k}
\end{table}
\noindent \textbf{Ablation of $\text{Top-}k$.}

We systematically examined the effects of various $\text{Top-}k$ strategies on model detection performance. As shown in Tab.~\ref{tab:ab_top-k}, setting $k$ to 25\% gave the best performance across multiple metrics, suggesting that this value provides a good balance between accuracy and information retention. The value of $k$ affects the model’s focus area; when $k$ is too small, the model lacks enough information, leading to a drop in performance.

Additional ablation experiment results are provided in the appendix, including more hyperparameter analysis, the impact of different augmentation information and training losses, and the effects of varying tampered area and subclass distributions on model performance.

\section{Conclusion}

This paper addresses the limitations of existing AIGC detection datasets, which largely focus on full-generated or object-level forgeries. We introduce BR-Gen, a high-quality dataset with 150,000 locally forged images, covering underrepresented stuff and background regions. To better detect subtle and spatially scattered forgeries, we propose NFA-ViT, a noise-guided transformer that amplifies forgery features across the image through attention modulation. Experimental results show that BR-Gen poses significant challenges to current methods, while NFA-ViT achieves strong and consistent performance. Our work provides a new foundation for advancing localized forgery detection in more diverse and realistic settings.

\section*{Acknowledgments}
This work was supported by National Key R\&D Program of China (No.2023YFB4502804), the National Science Fund for Distinguished Young Scholars (No.62025603), the National Natural Science Foundation of China (No. U22B2051, No. 62302411) and China Postdoctoral Science Foundation (No. 2023M732948).

\bibliography{aaai2026}

\newpage
\begin{center}
    \Large\bfseries Supplementary materials
\end{center}
\vspace{0em}  
\setcounter{section}{0}
\setcounter{subsection}{0}
\renewcommand{\thesection}{\Alph{section}}
\renewcommand{\thesubsection}{\thesection.\arabic{subsection}}

\section{Related Work}
\label{sec:relatedwork}

\subsection{Generation Datasets}
\noindent \textbf{Image Generation.} In recent years, with the rapid development of artificial intelligence and deep learning, Artificial Intelligence Generated Content (AIGC) has become widely used. Due to concerns about content security, the detection of generated images has gained increasing attention. Datasets~\cite{yan2024sanity,bird2024cifake,DBLP:conf/ccs/ShaLYZ23,zhu2023genimage,ojha2023towards,wang2023dire} containing both real and generated images have been organized for training and evaluating detection systems. Early datasets like CNNSpot ~\cite{wang2020cnn} collected fake images from various GAN architecture generators ~\cite{DBLP:conf/iclr/KarrasALL18, karras2019style, karras2020analyzing, DBLP:conf/iclr/BrockDS19, zhu2017unpaired, choi2018stargan,park2019semantic}. With the emergence of more advanced architectures like Diffusion Model ~\cite{ho2020denoising} and its variants ~\cite{dhariwal2021diffusion, nichol2021glide, nichol2021improved,rombach2022high,song2020denoising,liu2022pseudo, lu2022dpm,hertz2022prompt}, high-quality generated images have made discrimination more challenging. The later GenImage ~\cite{zhu2023genimage} provided a benchmark evaluation test with millions of images. Chameleon ~\cite{yan2024sanity} offered the most 'realistic' generated image test set. However, these datasets are mainly suitable for image-level detection tasks and fail to meet the requirements for local generation detection. Creating datasets for local generation tasks is more costly, and the pixel-level annotation process is more complex.

\noindent \textbf{Localized Image Generation.} Detecting generated or edited regions in images has been a longstanding challenge. Table ~\ref{tab:dataset_attribute} summarizes existing datasets, comparing their scale, data sources, generation techniques, and mask types. This includes recent generative tampering datasets like CocoGLIDE ~\cite{guillaro2023trufor}, IMD20 ~\cite{novozamsky2020imd2020}, AutoSplice ~\cite{jia2023autosplice}, TGIF ~\cite{mareen2024tgif}, and GRE ~\cite{sun2024rethinking}, all widely used and recognized in the field. For recent local generation datasets, we've identified a potential data bias in their construction process, which relies on object masks with clearly countable objects. These masks can be obtained directly from the COCO dataset ~\cite{lin2014microsoft} or through automatic segmentation using SAM ~\cite{kirillov2023segment}. Such masks neglect broader image regions, specifically the `stuff' category (sky, grassland, ground) and the `background' category (the inverse of object masks). Various generation detection models exhibit significantly reduced generalization performance on these two types due to this inherent bias.

\subsection{Generation Detection} 
\noindent \textbf{AIGC Detection.} The need for detecting generated images has been present since the emergence of deep learning. Early studies primarily focused on spatial domain features, such as color ~\cite{mccloskey2018detecting}, reflection ~\cite{o2012exposing}, and saturation ~\cite{mccloskey2019detecting}. As generative architectures advanced, CNNSpot ~\cite{wang2020cnn} demonstrated that image classifiers trained exclusively on ProGAN ~\cite{karras2017progressive} generators could generalize effectively to other unseen GAN architectures ~\cite{karras2019style, karras2020analyzing, DBLP:conf/iclr/BrockDS19, zhu2017unpaired} through carefully designed data augmentation and post-processing techniques. Recent approaches ~\cite{wang2023dire, qian2020thinking, liu2024forgery, ojha2023towards} have introduced novel strategies to improve generalization. F3Net ~\cite{qian2020thinking} investigates frequency differences between generated and real images, leveraging these variations for detection. DIRE ~\cite{wang2023dire} generates features by computing the difference between an image and its reconstruction using a pre-trained ADM ~\cite{dhariwal2021diffusion}, aiding the training of deep classifiers. FatFormer ~\cite{liu2024forgery} employs forgery-aware adapters that detect and integrate local forgery traces based on CLIP. Nonetheless, these methods focus solely on analyzing the entire image to determine authenticity, without identifying specific forged regions.

\noindent \textbf{Localized AIGC Detection.} Numerous methods~\cite{su2024can,zhu2024mesoscopic,guillaro2023trufor,liu2022pscc} have been proposed to identify forged areas in images. Wu \emph{et al}.~\cite{wu2019mantra} introduced ManTra-Net, which uses Long Short-Term Memory techniques to detect various forgery traces. MVSS-Net~\cite{dong2022mvss} employs a dual-stream CNN to extract noise features and incorporates a double attention mechanism to combine its outputs. Trufor~\cite{guillaro2023trufor} utilizes learned noise-sensitive patterns to identify generation traces. SparseViT~\cite{su2024can} addresses semantic inconsistencies in generated images by applying sparse attention for feature learning within sparse blocks, achieving state-of-the-art results. While these methods perform well in detecting "cheapfake" forgeries, they face challenges with complex mask types. We evaluate the performance of these models using our proposed evaluation framework.

\section{BR-Gen Generated Pipeline}

\subsection{Qwen2.5-VL Prompts}
\begin{figure}[]
    \centering
    \includegraphics[width=0.48 \textwidth]{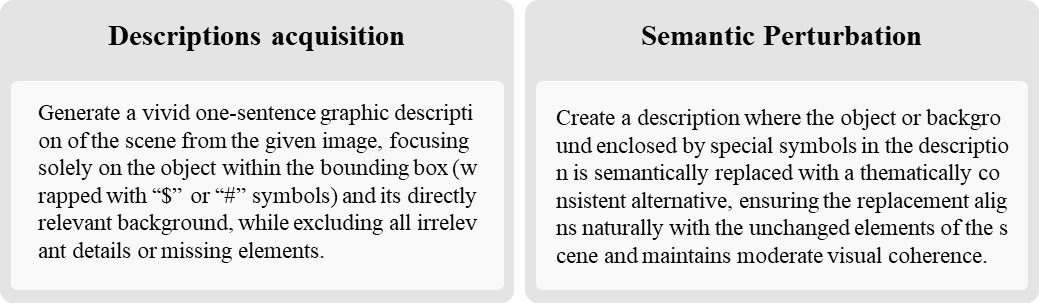}
    \caption{Qwen2.5-VL prompts.}
    \label{fig:prompt}
\end{figure}

We first present the prompt used in the \textbf{Perception} stage, as shown in Fig.~\ref{fig:prompt}. We employed Qwen2.5-VL in two phases to generate corresponding descriptions. The first stage extracts object and background perceptions from images to obtain visual descriptions. To enhance semantic richness and diversity, the second stage introduces probabilistic semantic perturbations, deriving modified descriptions from original images and their captions for subsequent diversified guidance.

\subsection{Localized Generation Pipeline}
\subsubsection{Preception.} 
The first critical step in the pipeline involves perceiving real images, which includes generating forged region masks and achieving a semantic understanding of the image content and tampered areas. 
We select the candidate categories by ``thing category'' and ``stuff category'' from COCO.
We employ GroundingDINO~\cite{liu2023grounding}, an open-set multi-modal object detection model to locate bounding boxes for those categories and select the target with the highest confidence.
Then SAM2~\cite{ravi2024sam2} will then convert these bounding boxes into their corresponding masks. 
SAM2 directly obtains masks for ``stuff'', while the ``background'' type is derived by inverting the masks of the ``thing''.
To mitigate category bias from overemphasizing specific categories, we manually control the number of all the objects is balanced.


For subsequent prompt-based inpainting methods, additional regional guiding prompts are required as input. We employ Qwen2.5-VL~\cite{DBLP:journals/corr/abs-2502-13923}, to recognize both global semantics of the image and forged regions. 
Specifically, we input both the original image and annotated images with bounding boxes into the model to obtain descriptions containing the edited target and the entire image.

To enhance relevance between generated content and original areas while increasing semantic diversity in generated content, we propose \textbf{Semantic Perturbation} to modify text descriptions related to generated content semantics. 
Specifically, annotated images and descriptions are re-input into Qwen2.5-VL. For ``stuff'', semantic replacement is performed on content enclosed in special symbols ``\#'' within descriptions while ensuring that replaced semantic information remains consistent with original areas (e.g., ``the blue sky'' $\to$ ``the starry sky''). For ``background'', text outside special symbols ``\$'' is replaced. 
To maintain consistency with original images while promoting semantic diversity, we set up semantically similar object lists to prevent significant semantic changes. 

\begin{figure*}[t]
    \centering
    \includegraphics[width=1 \textwidth]{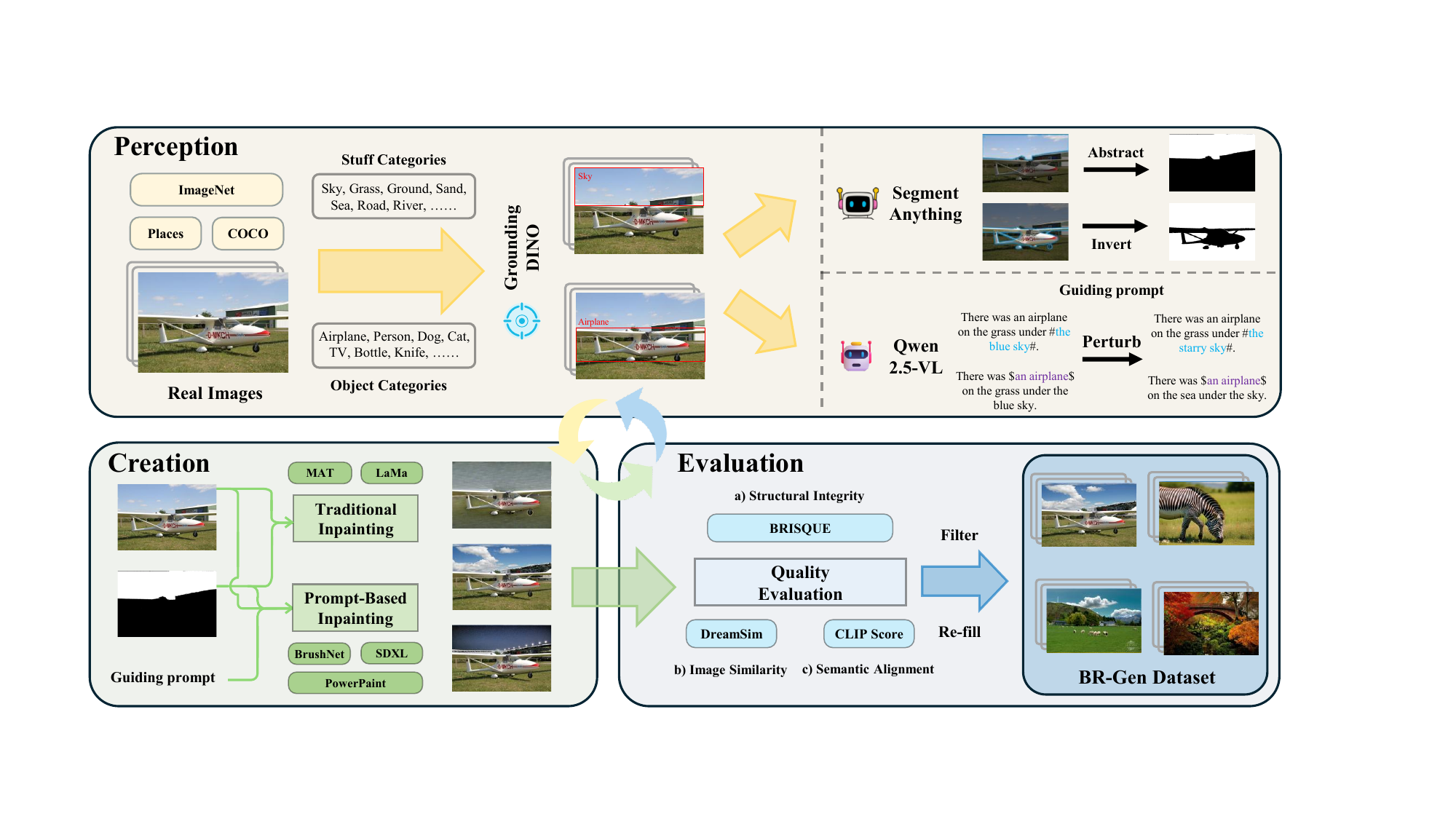}
    \caption{The automated pipeline for the BR-Gen dataset consists of three iterative stages: \textbf{Perception}, \textbf{Creation}, and \textbf{Evaluation}. These stages are applied to produce high-quality localized generation datasets through progressive refinement.}
    \label{fig:br-gen}
\end{figure*}

\subsubsection{Creation.} 
After all the required information for localized generation has been collected, including binary masks indicating areas to be forged and guiding prompts specifying the content for these regions, those data serves as detailed instructions for localized editing.

Previous studies~\cite{wang2020cnn} have examined that detection models demonstrate varying generalization performance when applied to data generated by different methods. 
To ensure diversity in the edited images within our BR-Gen dataset and provide a reliable benchmark for evaluating generalization, we employed five widely used and advanced inpainting methods to complete this process. 
These methods can be divided into two categories based on their architectures and generation approaches: traditional GAN-based inpainting methods: LaMa~\cite{suvorov2022resolution} and MAT~\cite{li2022mat}; and prompt-based inpainting methods built on Diffusion architectures: SDXL~\cite{podell2023sdxl}, BrushNet~\cite{ju2024brushnet}, and PowerPaint~\cite{zhuang2024task}. 

For traditional inpainting methods, only the original image and region mask are required as inputs to generate the corresponding inpainted image. For prompt-based inpainting methods, the original image, region mask, and guiding prompt must all be provided to produce the corresponding inpainted image.

\subsubsection{Evaluation.} 
To improve the quality of forged images, it is necessary to assess their quality and filter those that meet the required standards. Our evaluation process focuses on: (a) the structural integrity of the images; (b) the similarity between the generated images and the original images; and (c) the semantic alignment between prompt-based images and their corresponding captions.

First, we investigated methods to measure the structural integrity of individual images. BRISQUE~\cite{mittal2012no}, a no-reference metric designed to assess perceptual quality, was used for evaluation. Higher BRISQUE scores indicate lower perceptual quality, and images with scores above 60 were excluded. 
Second, we selected DreamSim~\cite{fu2023dreamsim} for evaluating the similarity before and after editing.
It integrates multiple foundational models~\cite{radford2021learning, ilharco_gabriel_2021_5143773, zhangdino} to evaluate both low-level and high-level similarity metrics, aligning closely with human perceptions. 
Additionally, we employed CLIP scores to ensure that image variations were consistent with the guiding prompts. Images with very low scores were removed.

Following this quality assessment process, each image was subjected to detection and filtering. To address data shortages caused by filtering, additional images were generated through iterations of the automated pipeline to increase dataset size. Ultimately, we created a high-quality BR-Gen Dataset using this automated workflow.

\begin{figure*}[t]
    \centering
    \includegraphics[width=1 \textwidth]{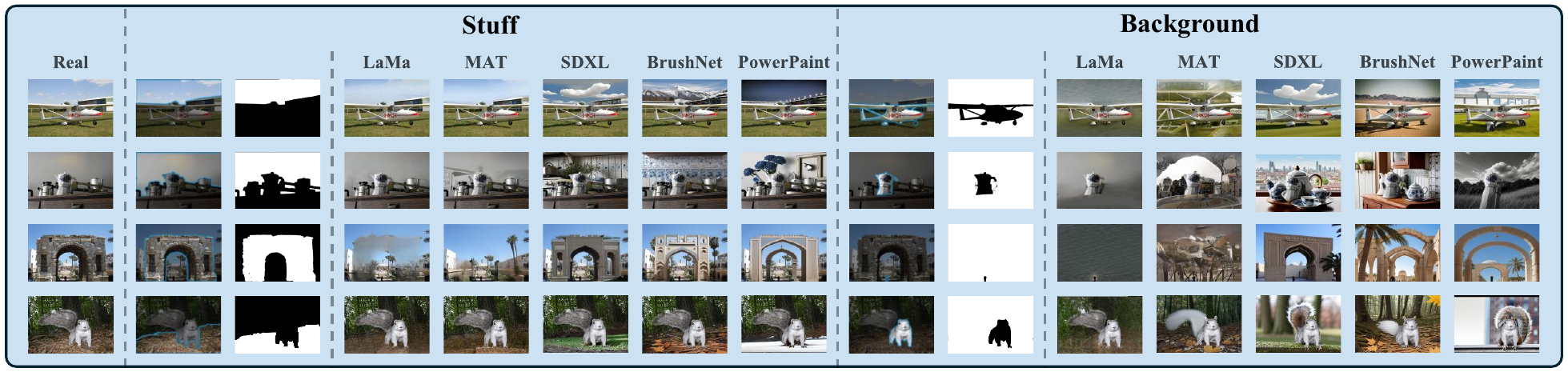}
    \caption{Example images from the BR-Gen dataset. Each row represents a pairing, with the first column displaying the real image, the second and third columns showing the region masks, and the fourth to eighth columns presenting the locally generated images. The two main groups illustrate the generative effects for Stuff and Background categories. The real source data comes from the open-source dataset Places~\cite{zhou2017places}, COCO~\cite{lin2014microsoft}, and ImageNet~\cite{deng2009imagenet}.}
    \label{fig:cases}
\end{figure*}

\subsection{Inpainting Methods}
To achieve localized image generation, we consider the most widely used inpainting models. Given an original image and a region mask, these models remove or edit objects in specified areas. Our approach incorporates diverse generative architectures spanning from conventional GANs to Diffusion Models, ensuring comprehensive coverage of mainstream and state-of-the-art methodologies.
\begin{itemize}
    \item \textbf{LaMa} (WACV'2022)\cite{suvorov2022resolution} is currently the most popular inpainting model, which has been deeply integrated into multiple applications to achieve image inpainting functionality. Many subsequent studies have also further expanded upon LaMa.

    \item \textbf{MAT} (CVPR'2022)\cite{li2022mat} is the most advanced image restoration model in current GAN architectures, taking only masked inputs and outputting the corresponding restored images.

    \item \textbf{SDXL} (ICLR'2024)\cite{podell2023sdxl} is a advanced high-resolution text-to-image generation model, with its core architecture built upon the Latent Diffusion Model. It achieves high-quality image synthesis through a two-stage generation process. SDXL-Inpainting~\cite{diffusers2023sdxl-inpainting} specializes in localized image editing tasks, enabling precise control of repair areas using masks and seamlessly integrating content generated from textual prompts with the original image.

    \item \textbf{BrushNet} (ECCV'2024)\cite{ju2024brushnet} is a diffusion-based text-guided image inpainting model that can be plug-and-play into any pre-trained diffusion model.Leveraging dense per-pixel control over the entire pre-trained model enhances its suitability for image inpainting tasks.

    \item \textbf{PowerPaint} (ECCV'2024)\cite{zhuang2024task}  is a high-quality versatile image inpainting model that supports text-guided object inpainting, object removal, shape-guided object insertion, and outpainting at the same time.
    
\end{itemize}

\subsection{Showcase}
To visually illustrate the effectiveness of our automated pipeline and the quality of the resulting dataset, we present several examples in Fig.~\ref{fig:cases}. These examples are sourced from various data origins, regional mask types, and inpainting methods. Each row in the figure is displayed in a paired format, consisting of the original image, region mask, and generated image.

\subsection{Mask Area}
\begin{figure}[]
    \centering
    \includegraphics[width=0.48 \textwidth]{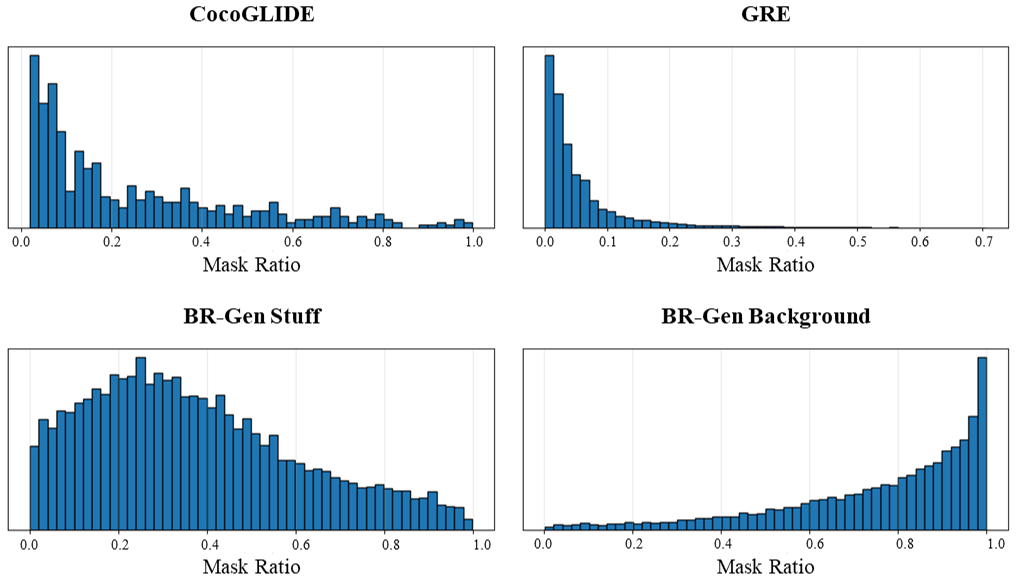}
    \caption{Mask area distribution.}
    \label{fig:area}
\end{figure}

In this analysis, we examined the distribution of masked regions in the BR-Gen dataset, where different mask types significantly alter the masked regions. We visualized two test datasets used in localized detection, CocoGLIDE~\cite{guillaro2023trufor} and GRE~\cite{sun2024rethinking}, and further categorized Stuff and Background from GR-gen as subcategories for visualization. The area distribution is shown in Fig.~\ref{fig:area}. It is observed that existing localized forgery datasets typically have mask areas concentrated in small regions due to prior knowledge of object masks. Our approach with Stuff and Background improves this limitation, enabling a more rigorous evaluation of the performance of current localized forgery detection models.

\subsection{Potential issues and analysis}
In Perception stage, the performance of base models like GroundingDINO and Qwen2.5-VL maybe influence the quality and diversity of generated forgeries. we taken several steps to mitigate these risks during the construction of BR-Gen.  To ensure quality, we implemented a multi-level filtering and  evaluation pipeline, which systematically removed low-quality or semantically inconsistent samples. A manual audit of 1,000 randomly sampled images from the final dataset showed that fewer than 1\% exhibited visible inconsistencies or annotation artifacts, indicating strong overall quality control.

Beyond filtering, we also intentionally diversified the source prompts and manipulation targets to cover rare object categories, uncommon spatial configurations, and varied scene types. In this process, we set up a list of similar categories (e.g., ``grass'' $\rightarrow$ ``ground'', ``sunny sky'' $\rightarrow$ ``blue sky'') during the semantic perturbation process to prevent the model from producing images with significant semantic deviation during perception and creation. Finally, We assessed semantic richness, confirming over 300 distinct semantics in the dataset, ensuring substantial diversity.

\section{NFA-ViT}

\subsection{Effectiveness of Noiseprint++}
\begin{figure}[]
    \centering
    \includegraphics[width=0.48 \textwidth]{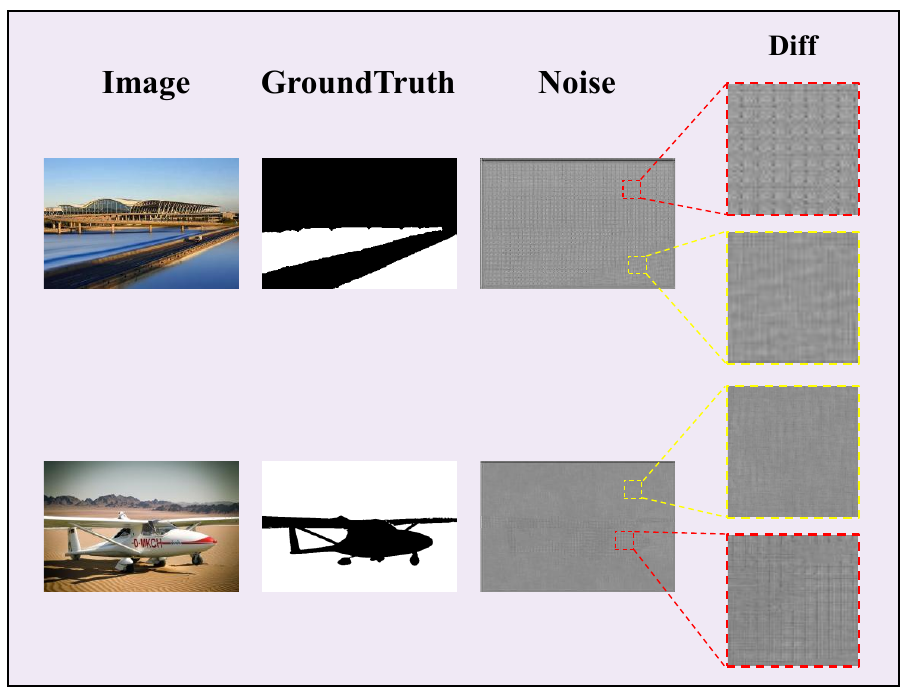}
    \caption{Noiseprint++ differential feature visualization for tampered images.}
    \label{fig:noise}
\end{figure}

The Noiseprint++ extractor is designed to capture device- and position-specific characteristics in real images through contrastive learning. To evaluate its applicability in distinguishing between generated and authentic image regions, we conducted a qualitative analysis based on noise visualization, as shown in Fig.~\ref{fig:noise}. The results demonstrate that Noiseprint++ features exhibit distinct patterns across different regions, indicating their potential as discriminative guidance for identifying generated content. Furthermore, by integrating both quantitative noise ablation studies and qualitative visual analysis presented in the main text, we validate the effectiveness and robustness of the proposed approach.

\subsection{Effectiveness of NAA module}
\begin{figure}[]
    \centering
    \includegraphics[width=0.48 \textwidth]{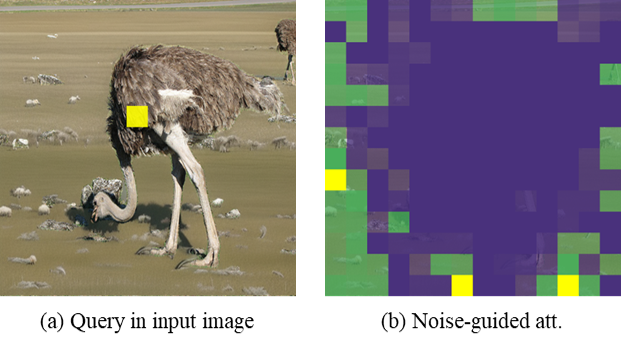}
    \caption{Visualization of attention distribution in NAA module.}
    \label{fig:attn}
\end{figure}

In the structural ablation study presented in paper, we quantitatively evaluated the contribution of the NAA module to overall performance. To further validate its functionality, we visualized the attention distribution of features within the NAA module, as shown in Fig.~\ref{fig:attn}. The results indicate that when the query involves objects, the model predominantly focuses on the background regions. This observation demonstrates the NAA module's ability to adaptively allocate attention and highlights its effectiveness in capturing contextually relevant features.

\begin{table*}[]
  \centering
  \resizebox{\textwidth}{!}{
    \begin{tabular}{@{}clccccccccc@{}}
    \toprule
     \multirow{2}{*}{\textbf{Task}} & \multirow{2}{*}{\textbf{Method}} &
     \multicolumn{3}{c}{\textbf{CoCoGLIDE}} & \multicolumn{2}{c}{\textbf{GRE}} & \multicolumn{2}{c}{\textbf{ForenSynths\_StyleGAN}} & \multicolumn{2}{c}{\textbf{Ojha\_Glide\_100\_27}} \\
        \cmidrule(lr){3-5}
        \cmidrule(lr){6-7}
        \cmidrule(lr){8-9}
        \cmidrule(lr){10-11}
     
   ~ & ~ & \multicolumn{1}{c}{\textbf{Gen. R@50}} & \multicolumn{1}{c}{\textbf{Real R@50}} & \multicolumn{1}{c}{\textbf{IoU}} & \multicolumn{1}{c}{\textbf{Gen. R@50}} & \multicolumn{1}{c}{\textbf{IoU}} & \multicolumn{1}{c}{\textbf{Gen. R@50}} & \multicolumn{1}{c}{\textbf{Real R@50}} & \multicolumn{1}{c}{\textbf{Gen. R@50}} & \multicolumn{1}{c}{\textbf{Real R@50}} \\ \midrule
    \multirow{2}{*}{\textbf{Localized Detection}} 
    & Trufor~\cite{guillaro2023trufor} & 0.762 & 0.952 & 0.781 & 0.787 & 0.677 & 0.916 & 0.946 & 0.911 & 0.955  \\
    & Sparse ViT~\cite{su2024can} & \underline{0.876} & 0.989 & \underline{0.833} & \underline{0.852} & \underline{0.720} & 0.945 & \textbf{0.991} & 0.947 & \underline{0.992} \\ 
    \midrule
    
    \multirow{2}{*}{\textbf{AIGC Detection}}
    & NPR~\cite{tan2024rethinking} & 0.833 & 0.957 & - & 0.810 & - & 0.942 & 0.945 & 0.932 & 0.952 \\
    & FatFormer~\cite{liu2024forgery} & 0.859 & \textbf{0.992} & - & 0.843 & - & \underline{0.967} & \underline{0.990} & \underline{0.955} & 0.991 \\
    \midrule
    \rowcolor{green!12}
& \textbf{NFA-ViT} ~(ours) & \textbf{0.884} & \underline{0.990} & \textbf{0.856} & \textbf{0.865} & \textbf{0.774} & \textbf{0.972} & \textbf{0.991} & \textbf{0.962} & \textbf{0.995} \\
    
    \bottomrule
    \end{tabular}
  }

    \caption{The generalization results on existing datasets.}
      \label{tab:existing_dataset}

\end{table*}

\begin{table*}[]
  \centering
  \resizebox{\textwidth}{!}{
    \begin{tabular}{@{}clccccccccc@{}}
    \toprule
     \multirow{2}{*}{\textbf{Task}} & \multirow{2}{*}{\textbf{Method}} &
     \multicolumn{2}{c}{\textbf{GAN $\to$ Diffusion}} & \multicolumn{2}{c}{\textbf{Diffusion $\to$ GAN}} & \multicolumn{2}{c}{\textbf{Background $\to$ Stuff}} & \multicolumn{2}{c}{\textbf{Stuff $\to$ Background}} \\
        \cmidrule(lr){3-4}
        \cmidrule(lr){5-6}
        \cmidrule(lr){7-8}
        \cmidrule(lr){9-10}
     
   ~ & ~ & \multicolumn{1}{c}{\textbf{Gen. R@50}} & \multicolumn{1}{c}{\textbf{Real R@50}} & \multicolumn{1}{c}{\textbf{Gen. R@50}} & \multicolumn{1}{c}{\textbf{Real R@50}} & \multicolumn{1}{c}{\textbf{Gen. R@50}} & \multicolumn{1}{c}{\textbf{Real R@50}} & \multicolumn{1}{c}{\textbf{Gen. R@50}} & \multicolumn{1}{c}{\textbf{Real R@50}} \\ \midrule
    \multirow{2}{*}{\textbf{Localized Detection}} 
    & Trufor~\cite{guillaro2023trufor} & 0.206 & 0.964 & 0.405 & 0.962 & 0.709 & 0.904 & 0.673 & 0.932  \\
    & SparseViT~\cite{su2024can} & 0.373 & \underline{0.970} & 0.605 & 0.959 & \textbf{0.842} & 0.967 & \underline{0.883} & \underline{0.959} \\ 
    \midrule
    
    \multirow{2}{*}{\textbf{AIGC Detection}}
    & NPR~\cite{tan2024rethinking} & 0.225 & 0.962 & 0.468 & \underline{0.968} & 0.743 & 0.932 & 0.755 & 0.920 \\
    & FatFormer~\cite{liu2024forgery} & \underline{0.412} & 0.967 & \underline{0.725} & 0.956 & 0.795 & \underline{0.971} & 0.847 & 0.955 \\
    \midrule
    \rowcolor{green!12}
    ~ & ~~\textbf{NFA-ViT} ~(ours) & \textbf{0.466} & \textbf{0.980} & \textbf{0.820} & \textbf{0.973} & \underline{0.841} & \textbf{0.982} & \textbf{0.908} & \textbf{0.970} \\
    
    \bottomrule
    \end{tabular}
  }

    \caption{Cross-type in terms of R@50 on different type subsets.}
      \label{tab:cross_type}

\end{table*}

\begin{table}[]
\centering
\resizebox{.48\textwidth}{!}{
\begin{tabular}{l ccc}
\toprule
          & \textbf{\makecell{All\\Mani R@50}} & \textbf{\makecell{Sky\\Mani R@50}} & \textbf{\makecell{Wall\\Mani R@50}} \\
\midrule
SparseViT & 0.911           & 0.905           & 0.899            \\
\rowcolor{green!12}
NFA-ViT   & 0.953           & 0.957           & 0.950            \\
\bottomrule
\end{tabular}}
\caption{Detection results in the subclass categories.}
\label{tab:subclass}
\end{table}

\begin{table*}[]
  \centering
  \resizebox{1\textwidth}{!}{
    \begin{tabular}{lccccccc}
    \toprule
      \multirow{2}{*}{\textbf{Method}} &\multirow{2}{*}{\textbf{Original}}&
     \multicolumn{2}{c}{\textbf{Gaussian Noise}} & \multicolumn{2}{c}{\textbf{Gaussian Blur}} & \multicolumn{2}{c}{\textbf{JPEG Compression}} \\
        \cmidrule(lr){3-4}
        \cmidrule(lr){5-6}
        \cmidrule(lr){7-8}
     
 ~ &~& $\sigma=1$ & $\sigma=3$ & $\sigma=1$ &$\sigma=3$ & $q=.95$ & $q=.75$ \\
 
 \midrule
     SparseViT~\cite{su2024can} &91.13& 90.96(-0.17) & 87.92(-3.21)& 87.37(-3.76) & 85.92(-5.21) & 90.11(-1.02)& 88.82(-2.31)  \\
    AIDE~\cite{yan2024sanity} & 94.14 & 94.11(-0.03) & 92.13(-2.01) & 91.44(-2.70) & 88.77(-5.37) & 92.98(-1.16) & 91.71(-2.43)  \\
    \rowcolor{green!12}
    \textbf{NFA-ViT} ~(ours) & \textbf{95.32} & \textbf{95.24(-0.08)} & \textbf{92.75(-2.57)} & \textbf{92.66(-2.66)} & \textbf{90.23(-5.09)} & \textbf{95.27(-0.05)} & \textbf{93.68(-1.64)} \\
    
    \bottomrule
    \end{tabular}}

    \caption{Robustness to unseen perturbations. We present R@50 for two distinct degradation levels across three types of perturbations.}
      \label{tab:robust}

\end{table*}

\begin{table}[]
  
  \centering
  \resizebox{0.44\textwidth}{!}{
  
    \begin{tabular}{@{}cccc@{}}
    \toprule
      & \textbf{Gen. R@50} & \textbf{Real R@50} & \textbf{IoU} \\
     
    \midrule
    10\% & 0.945 & 0.989 & 0.887 \\
    \rowcolor{green!12}
    25\% & \textbf{0.953} & 0.992 & \textbf{0.907} \\ 
    50\% & 0.947 & \textbf{0.993} & 0.897 \\ 
    \bottomrule
    \end{tabular}}
  \caption{Different value of $\text{Top-}k$ in Noise-Guided Amplification Attention.}
  \label{tab:ab_top-k}
\end{table}
\begin{table}[]

  \resizebox{.47\textwidth}{!}{
  \centering
  
    \begin{tabular}{@{}lccccc@{}}
    \toprule
      \textbf{Method}& $<20\%$ & $<40\%$ &$<60\%$ &$<80\%$ & $<100\%$ \\
     
    \midrule
    TruFor & 0.899 &	0.897& 	0.895 &	0.891 	&0.887  \\
    SparseViT & 0.917 	&0.913& 	0.910 	&0.906& 	0.904  \\ 
    \midrule
    NPR & 0.882 &	0.896& 	0.900 &	0.902 	&0.906  \\ 
    FatFormer & 0.920 	&0.927& 	0.930 	&0.936& 	0.941  \\ 
    \midrule
    \rowcolor{green!12}
    \textbf{NFA-ViT}(ours) & \textbf{0.965} &	\textbf{0.960}& 	\textbf{0.954} &	\textbf{0.945} &	\textbf{0.948}  \\ 
    \bottomrule
    \end{tabular}}
  \caption{Value of the generated image R@50 under different mask area distributions.}
  \label{tab:ab_area}
\end{table}

\begin{table*}[]
\centering
\resizebox{1\textwidth}{!}{
\begin{tabular}{l ccccc}
\toprule
\textbf{Method} & \textbf{Size} & \textbf{Parameter(MB)} & \textbf{FLOPs(Gflops)} & \textbf{Inference Time(ms)} & \textbf{\makecell{Inference Memory\\Usage(GB)}} \\
\midrule
ManTraNet      & 256x256 & 3.9   & 274    & 42.02 & 7.0 \\
SparseViT      & 512x512 & 50.3  & 46.2   & 26.67 & 2.2 \\
MVSS           & 512x512 & 147   & 167    & 37.37 & 3.4 \\
TruFor         & 512x512 & 68.7  & 230.1  & 47.02 & 3.0 \\
\rowcolor{green!12}
Ours(NFA-ViT) & 512x512 & 60.25 & 156.68 & 40.33 & 2.6 \\
\bottomrule
\end{tabular}}
\caption{Comparison with the State-of-the-Art on Parameter and Inference consumption.}
\label{tab:model_info}
\end{table*}

\section{Experimental details}
\subsection{Detectors}

We selected a total of 11 representative AI-generated image detection methods as benchmarks for comparison in both AIGC detection and localized detection, covering models from early to recent stages and ranging from the most commonly used to the best-performing ones in the two tasks.

\begin{itemize}

    \item \textbf{LGrad} (CVPR'2023)\cite{tan2023learning} employs gradients computed by a pretrained CNN model to present the generalized artifacts for classification.

    \item \textbf{DIRE} (ICCV'2023)\cite{wang2023dire} observes obvious differences in discrepancies between images and their reconstruction by DMs and uses this feature to train a classifier.

    \item \textbf{FreqNet} (AAAI'2024)\cite{DBLP:conf/aaai/Tan0WGLW24}  introduces a lightweight CNN classifier that leverages frequency domain learning to enhance the generalizability of deepfake detection. 

    \item \textbf{NPR} (CVPR'2024)\cite{tan2024rethinking} contributes to the architectures of CNN-based generators and demonstrates that the up-sampling operator can generate generalized forgery artifacts that extend beyond mere frequency-based artifacts.

    \item \textbf{FatFormer} (CVPR'2024)\cite{liu2024forgery} introduces a forgery-aware adaptive transformer that integrates dual-domain analysis and vision-language alignment. It employs contrastive language-guided supervision to enhance generalization.

    \item \textbf{ManTraNet} (CVPR'2019)\cite{wu2019mantra} proposes an end-to-end fully convolutional network for detecting and localizing diverse image manipulations and demonstrates strong generalizability across unseen manipulation types through its anomaly-driven approach.

    \item \textbf{MVSS-Net} (TPAMI'2022) \cite{dong2022mvss} proposes multi-view multi-scale supervised networks for image manipulation detection, combining edge boundary artifacts and noise inconsistency through dual-branch architecture.

    \item \textbf{PSCC-Net} (TCSVT'2021) \citep{liu2022pscc} proposes a dual-path architecture with top-down feature extraction and bottom-up progressive mask estimation, incorporating Spatio-Channel Correlation Modules (SCCM) to capture holistic manipulation traces. 

    \item \textbf{TruFor} (CVPR'2023) \cite{guillaro2023trufor} proposes a transformer-based forensic framework that integrates multi-view clues from RGB content and learned noise fingerprints to detect image forgeries through anomaly detection. 

    \item \textbf{SparseViT} (AAAI'2025) \cite{su2024can} proposes a sparse self-attention mechanism that replaces dense global attention in Vision Transformers, enabling adaptive extraction of manipulation-sensitive non-semantic features through discrete patch interactions.
    \item \textbf{AIDE} (ICLR'2025) \cite{yan2024sanity} utilizes multiple experts to simultaneously extract visual artifacts and noise patterns, capture high-level semantics of images, and calculate low-level patch features.
\end{itemize}

\subsection{More Results}
We have added more comparative and ablation experimental results in this section, such as the cross-type testing in BR-Gen, the robustness evaluation, the hyperparameter $k$ in NAA module, the performance under the difference tampered area distribution, the use of different auxiliary information, and the way to obtain classification results (using additional classification headers or probabilistic information using segmentation plots).

\noindent \textbf{Generalization on Existing Datasets.}
We further evaluated the generalization ability of our models on multiple existing datasets~\cite{ojha2023towards,wang2020cnn,guillaro2023trufor,sun2024rethinking}. The models trained on BR-Gen were tested on these datasets, as shown in Tab.~\ref{tab:existing_dataset}. For the local AIGC detection dataset, we tested both the classic CocoGLIDE~\cite{guillaro2023trufor} and the largest GRE~\cite{sun2024rethinking}. In terms of localization performance, NFA-ViT outperformed SparseViT, confirming NFA-ViT's advantage in localization ability. In the AI-generated detection dataset, we used a subset called StyleGAN from ForenSynths~\cite{wang2020cnn} and Glide\_100\_27 by Ojha~\cite{ojha2023towards} to meet the testing needs for both GAN and diffusion architectures. The experimental results showed that models trained on BR-Gen generalize well to other detection datasets.

\noindent \textbf{Cross-type testing.}
As shown in Tab.~\ref{tab:cross_type}, we evaluated the transferability of various methods across different types to assess the generalization performance of the models in cross-style and cross-architecture settings. In ``GAN $\to$ Diffusion,'' all methods showed a clear drop in Gen. R@50, highlighting a large gap between data quality and generative architecture, which makes generalization more difficult. In the tests for Background and Stuff, the performance of ``Stuff $\to$ Background'' was better. Dataset analysis showed that Background in some cases also includes Stuff, which explains the differences in generalization between the two.

\noindent \textbf{Results in subclass categories.}
To investigate whether the model exhibits bias toward specific semantic subcategories (e.g., sky, grass) on the BR-Gen dataset, we conducted a fine-grained performance analysis across different manipulation categories. Specifically, we semantically categorized the tampered regions in each image and computed the tampering recall rate for each subclass, as summarized in Table~\ref{tab:subclass}.
The results indicate slight variations in detection performance across different categories. For instance, the sky category, characterized by its large spatial extent and distinct visual appearance, is relatively easier for most models to detect. In contrast, wall regions often exhibit weak texture contrast, leading to a notable performance drop in models such as SparseViT.
Notably, NFA-ViT demonstrates consistent and superior performance across all subcategories , maintaining high recall rates even in visually subtle or less discriminative regions. This suggests that our method is robust to semantic category variations and less prone to subclass bias.

\noindent \textbf{Robust Evaluation.} 
To assess the robustness of our model against various perturbations, we followed the evaluation protocol used in prior works~\cite{DBLP:conf/iccv/ZhengB0ZW21}, which involves three common types of image degradation: Gaussian Noise, Gaussian Blur, and JPEG Compression. For each degradation type, we applied varying levels of perturbation to evaluate model performance under increasing degrees of distortion. The results are summarized in Tab.~\ref{tab:robust}.
As shown in the table, NFA-ViT exhibits the smallest performance degradation compared to other state-of-the-art models across all degradation settings. Notably, despite our method's reliance on noise features, it still achieves superior performance under Gaussian Noise perturbations. These results demonstrate that NFA-ViT offers enhanced robustness to various types of image distortions. 

\noindent \textbf{Comparison of inference consumption.} Computational efficiency is critical for real-world deployment. we provide a detailed comparison of model parameters, FLOPs, inference time, and memory usage across representative methods, as summarized in the Tab.~\ref{tab:model_info}. The main results are also cited from~\cite{su2024can}. NFA-ViT achieves a favorable trade-off between performance and efficiency. Despite its dual-branch design and enhanced attention mechanisms, its model size and inference time remain comparable to, or better than, existing high-performance baselines such as TruFor and MVSS. NFA-ViT completes inference within 40.33 ms per image and requires only 2.6 GB of memory, making it lightweight enough for deployment on various edge devices with limited computational resources. This demonstrates its strong potential for practical applications.

\noindent \textbf{Ablation of $\text{Top-}k$.}
We systematically examined the effects of various $\text{Top-}k$ strategies on model detection performance. As shown in Tab.~\ref{tab:ab_top-k}, setting $k$ to 25\% gave the best performance across multiple metrics, suggesting that this value provides a good balance between accuracy and information retention. The value of $k$ affects the model’s focus area; when $k$ is too small, the model lacks enough information, leading to a drop in performance.

\noindent \textbf{Performance under different mask areas.}
To clearly analyze whether different models show bias in detecting various mask areas, we compared the performance changes of several models across different mask levels on BR-Gen. The results are shown in Tab.~\ref{tab:ab_area}. When the forged area is too small ($<$20\%), real features dominate, and methods that rely on global information without focusing on forged regions perform poorly. Our NFA-ViT improved by \textbf{4.5\%} in this range. As the forged area increases, local AIGC detection models generally show a drop in performance. However, AIGC detection models improve as the area grows, matching their task focus. Even with large forged areas, our NFA-ViT still achieved strong performance.

\begin{table}[]

  \setlength{\tabcolsep}{5.5mm}
  \centering
  
    \begin{tabular}{@{}cccc@{}}
    \toprule
      & \textbf{Gen. R@50} & \textbf{Real R@50} & \textbf{IoU} \\
     
    \midrule
    None & 0.926 & 0.986 & 0.841 \\
    \rowcolor{green!12}
    Noise & \textbf{0.953} & \textbf{0.992} & \textbf{0.907} \\ 
    DCT & 0.918 & 0.973 & 0.798 \\ 
    Sober & 0.940 & 0.987 & 0.858 \\ 
    Grad & 0.934 & 0.989 & 0.846 \\ 
    \bottomrule
    \end{tabular}
    \caption{Different augment information in NFA-ViT.}
  \label{tab:ab_aug_info}
\end{table}

\noindent \textbf{Ablation of Augment Info.}
We conducted ablation experiments on various conditional inputs in NFA-ViT, as shown in Tab.~\ref{tab:ab_aug_info}. The results show that adding auxiliary information clearly improves model performance. Notably, using noise as an auxiliary input gives the highest boost in detection accuracy, outperforming other types of conditional inputs such as DCT and Sobel images. Further analysis shows that while DCT captures image frequency information, its global nature limits its ability to guide local areas, leading to a clear drop in performance. In contrast, noise works as an invisible underlying feature of the image, making it useful for regional guidance due to the differences between generated and real regions.

\begin{table}[]

  \centering
  \setlength{\tabcolsep}{2.5mm}
  
    \begin{tabular}{@{}cccc@{}}
    \toprule
      & \textbf{Gen. R@50} & \textbf{Real R@50} & \textbf{IoU} \\
     
    \midrule
    Seg Loss & 0.930 & 0.982 & 0.901 \\
    \rowcolor{green!12}
    Seg Loss + Cls Loss & \textbf{0.953} & \textbf{0.992} & \textbf{0.907} \\ 
    \bottomrule
    \end{tabular}
    \caption{Ablation of Loss Usage.}
  \label{tab:usage_loss}
\end{table}
\noindent \textbf{Loss Function Analysis.} As shown in Tab.~\ref{tab:usage_loss}, we conducted comparative experiments on various loss function designs. When using only segmentation loss (Seg loss) to generate region masks, the output was considered correct only if the entire mask was zero. In the original setups of MantraNet and SparseViT, only Seg loss was used, and we followed this setup to evaluate their classification performance. The results show that even with Seg loss alone, the model can perform detection. However, adding classification loss (Cls loss) further improves model performance. This improvement comes mainly from the classification task's ability to capture global semantic information from images. When using only Seg loss, the model focuses on boundary features at the pixel level, which may be limited by region shape or unstable mask edges. By adding Cls loss, the model learns image-level forgery patterns during training, leading to stronger feature learning during optimization. This not only improves the accuracy of detecting forged regions but also helps the model generalize better across different image content.

\noindent \textbf{Failure Cases}
We conducted a systematic analysis of failure cases in NFA-ViT and identified the following patterns and conclusions:
\begin{enumerate}
\item When the boundary between the forged and authentic regions is visually ambiguous (e.g., in scenes involving skies or oceans), the model often fails to accurately localize the tampered areas.
\item In complex natural scenes such as forests, the irregular structures may lead to false positives, where genuine content is erroneously flagged as manipulated.
\item A majority of the failure cases involve images generated by advanced diffusion models such as SDXL, indicating the increasing challenge of detecting high-quality synthetic content.
\end{enumerate}

\section{Limitation and Future Work}
One current limitation of dataset construction pipeline is its reliance on existing tool models. The quality and diversity of the generated images are inherently constrained by the capabilities and limitations of these models. Another challenge lies in the use of widely adopted quality evaluation metrics, which may introduce bias and often do not fully align with human perception. This gap can affect the accuracy of dataset quality assessment in real-world applications.

In the future, we plan to integrate various types of image forgery into a unified framework, aiming to construct a large-scale, comprehensive dataset for forgery detection. Although our current work primarily focuses on detecting and localizing scene tampering, the proposed methodology has the potential to be extended to related tasks such as document tampered detection, and Deepfake detection. Moreover, we hope that our approach can inspire future research on more effective feature amplify strategies and adaptive learning mechanisms for tampered analysis.

\end{document}